\documentclass[10pt,twocolumn,letterpaper]{article}

\usepackage{iccv}
\usepackage{times}
\usepackage{epsfig}
\usepackage{graphicx}
\usepackage{amsmath}
\usepackage{amssymb}
\usepackage{booktabs}
\usepackage{multirow}
\usepackage{color}
\usepackage{bbm}
\usepackage{tabu}
\usepackage{threeparttable}
\usepackage{algorithmicx}
\usepackage{algpseudocode}
\usepackage{algorithm}
\usepackage{colortbl}
\usepackage[table]{xcolor}

\usepackage[pagebackref=true,breaklinks=true,letterpaper=true,colorlinks,bookmarks=false]{hyperref}

\iccvfinalcopy 

\def\iccvPaperID{4435} 

\ificcvfinal\fi

\begin{document}

\title{AirCache: Activating Inter-modal Relevancy KV Cache Compression for Efficient Large Vision-Language Model Inference}

\author{Kai Huang, Hao Zou, Bochen Wang, Ye Xi, Zhen Xie, Hao Wang\\
Alibaba Group}


\maketitle
\ificcvfinal\thispagestyle{empty}\fi

\begin{abstract}
   Recent advancements in Large Visual Language Models (LVLMs) have gained significant attention due to their remarkable reasoning capabilities and proficiency in generalization. However, processing a large number of visual tokens and generating long-context outputs impose substantial computational overhead, leading to excessive demands for key-value (KV) cache. To address this critical bottleneck, we propose AirCache, a novel KV cache compression method aimed at  accelerating LVLMs inference. This work systematically investigates the correlations between visual and textual tokens within the attention mechanisms of LVLMs. Our empirical analysis reveals considerable redundancy in cached visual tokens, wherein strategically eliminating these tokens preserves model performance while significantly accelerating context generation. Inspired by these findings, we introduce an elite observation window for assessing the importance of visual components in the KV cache, focusing on stable inter-modal relevancy modeling with enhanced multi-perspective consistency. Additionally, we develop an adaptive layer-wise budget allocation strategy that capitalizes on the strength and skewness of token importance distribution, showcasing superior efficiency compared to uniform allocation. Comprehensive evaluations across multiple LVLMs and benchmarks demonstrate that our method achieves comparable performance to the full cache while retaining only 10\% of visual KV cache, thereby reducing decoding latency by 29\% to 66\% across various batch size and prompt length of inputs. Notably, as cache retention rates decrease, our method exhibits increasing performance advantages over existing approaches.
   
\end{abstract}

\section{Introduction}
\label{sec:intro}
The past few years have witnessed remarkable advancements in Large Visual-Language Models (LVLMs) \cite{bai2023qwen, chen2024internvl, li2024llava, liu2024improved, liu2023visual, tong2025cambrian, wang2024qwen2, wu2024deepseek}, both in research and practical applications. Although LVLMs exhibit increasingly advanced visual processing capabilities, such as handling high-resolution, multiple images, and video sequences, these improvements entail substantial computational costs. The exponential increase in visual tokens, coupled with the demands of long-context generation tasks, results in unsustainable memory overhead from key-value (KV) cache storage \cite{pope2023efficiently}. This not only significantly increases GPU memory consumption but also severely degrades computational efficiency due to heightened memory bandwidth contention.
\begin{figure}[t!]
    \centering
    \setlength{\abovecaptionskip}{0cm}
    \setlength{\belowcaptionskip}{-0.5cm}
	\includegraphics[width=0.48\textwidth]{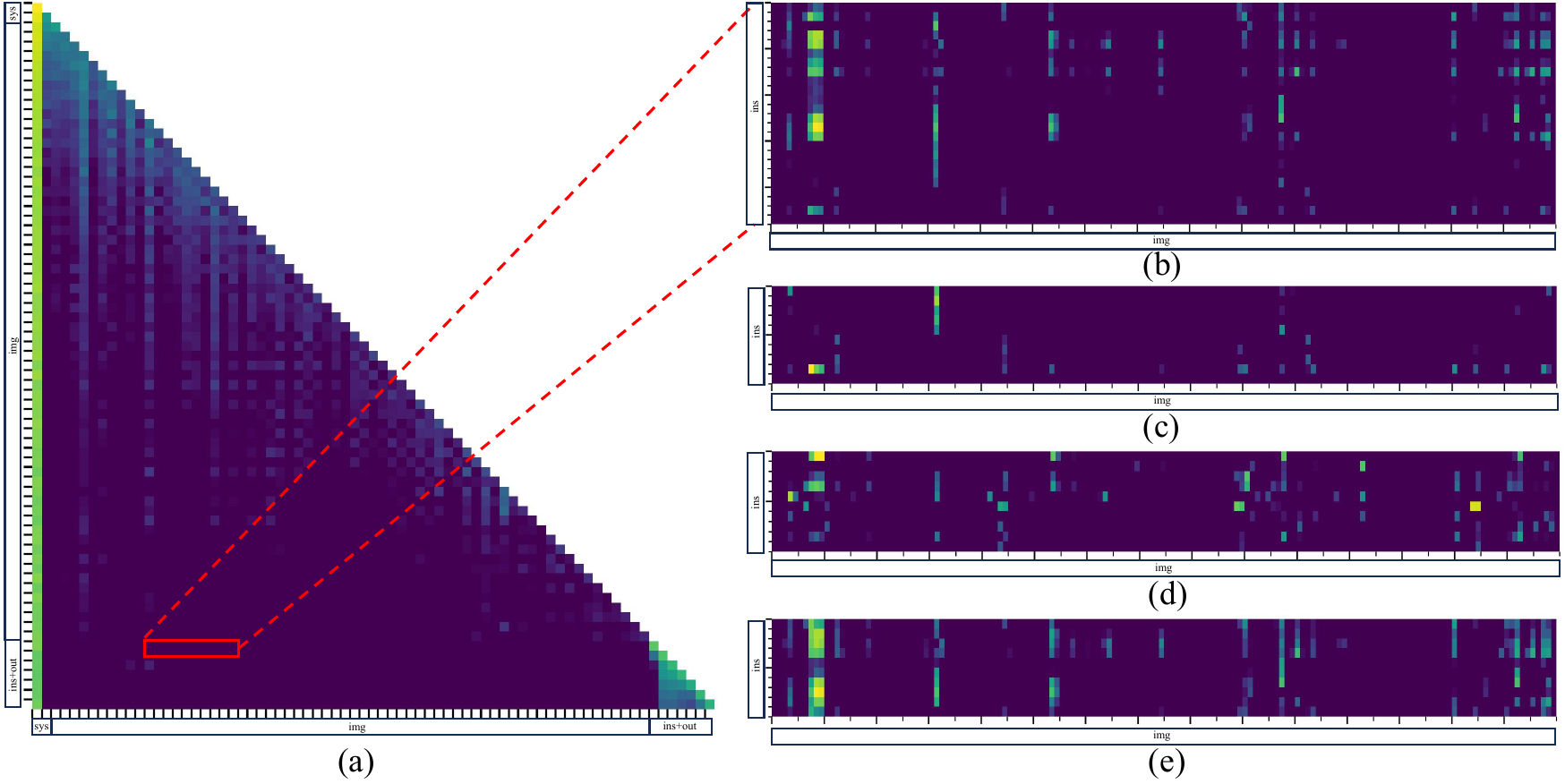}
	\caption{Illustration of our motivation. (a) An example of attention maps during the decoding process of LLaVA-OV-7B \cite{li2024llava}. (b) Partial visual attention maps obtained using all text tokens as the observation window. (c) Using the last 16 text tokens as the observation window. (d) Using the last 16 visual tokens as the observation window. (e) Using the proposed method as the observation window, demonstrating more consistent attention distribution.}
    \vspace{-10pt}
 	\label{fig1_motivation}
\end{figure}

Recent approaches \cite{chen2024image, huang2024ivtp, jiang2023bus, li2024snapkv, zhang2024cls, zhang2024h2o} aim to prune visual tokens during inference, which can be categorized into token pruning and KV cache compression. Token pruning \cite{he2024zipvl, huang2024ivtp, jiang2023bus, li2024tokenpacker, li2024llama, zhang2024cls} involves reducing visual tokens during the prefill phase, which decreases both the number of tokens processed in subsequent transformer layers and the tokens stored in the KV cache. Although it significantly enhances inference speed, the aggressive elimination of visual tokens during prefill leads to a substantial loss of visual information, severely degrading model performance. KV cache compression \cite{li2024snapkv, liu2025minicache, liu2024efficient, wang2024prefixkv, zhang2024h2o} reduces token solely during the decoding phase by pruning stored data in the KV cache. This technique has been extensively validated in LLMs. Since all tokens undergo a complete forward pass and the causal attention mechanism in LLMs differentiates the importance distribution among tokens, selectively deleting certain tokens has minimal impact on model performance. These methods typically assess the importance of visual tokens by evaluating their strength or sparsity within the attention maps computed under the causal attention mechanism. Given the strict unidirectionality of causal attention, using cross-modality attention scores between instruction tokens and visual tokens provides a more comprehensive evaluation of visual token importance than using full attention scores. Moreover, we observe significant variations in the distribution of visual token strength across different layers, indicating that uniformly allocating the budget is not the optimal strategy.

To address the aforementioned issues, we propose AirCache, a novel KV cache compression method designed for efficient LVLMs inference. AirCache comprises two main components: visual token importance scoring and layer-wise KV cache budget allocation. To effectively evaluate the importance of visual tokens in the KV cache, we introduce an elite observation window that selects critical instruction tokens by leveraging self-attention scores among instruction tokens rather than using all or continuous local instruction tokens. As illustrated in Figure \ref{fig1_motivation}, compared to mainstream methods that assess visual token importance based on observation windows guided by all or partial text instruction tokens, the elite observation window demonstrates stronger consistency. Most text tokens within the elite observation window tend to provide more similar evaluations for the same visual token, thereby enhancing the effectiveness and stability of subsequent voting-based ranking. We further propose to quantify the compression budget for different layers from two perspectives. The first aspect is the emphasis a layer places on visual information, measured by the sum of attention allocated to all visual tokens, as well as the strength of importance score distribution. Additionally, analysis of attention distributions across layers reveals a distinct head effect, where a few visual tokens receive high importance scores, while the majority exhibit mediocre and average importance characteristics. Our experimental results show that preserving only 10\% of the important visual KV cache results in less than 1\% average model performance degradation, corroborating the significant head effect in the importance distribution of visual tokens. Combining these two aspects, we allocate the budget for visual KV cache based on both the strength and skewness of importance scores distributions.

Our primary contributions include: (1) We conduct a comprehensive and detailed exploration of the differences in inter-modal relevancy patterns based on attention interaction in evaluating visual KV cache importance. We introduce an elite observation window with carefully selected key textural instruction tokens to achieve more effective and stable importance evaluations. (2) We propose to quantify the differentiated compression budget allocation across different layers using the strength and skewness of the importance scores distribution, based on the prominence of visual information and the efficacy of attention allocation. (3) Experimental results on widely used LVLMs and benchmark datasets demonstrate that the proposed method achieves superior performance compared to other existing methods.

\section{Related Work}
\label{sec:relatedwork}
{\bf Visual Token Pruning.} Visual token pruning \cite{cha2024honeybee, chen2024image, hu2024matryoshka, huang2024ivtp, li2024llama, zhang2024cls} accelerates model inference by reducing visual tokens during the prefill phase. BLIP-2 \cite{li2023blip} and Qwen-VL \cite{bai2023qwen} employ cross-attention with learnable query embeddings to aggregate input visual tokens, while HoneyBee \cite{cha2024honeybee} and MobileVLM \cite{chu2023mobilevlm} use convolutional neural networks. These approaches necessitate model retraining, which significantly increases migration costs. In contrast, methods like \cite{chen2024image, zhang2024cls} introduce no extra parameters but leverage attention maps to score and evict visual tokens based on importance. IVTP \cite{huang2024ivtp} implements two-stage pruning by deleting tokens in both Vision Transformer (ViT) and LLM, enabling text-guided pruning through instruction integration. However, it is important to note that visual token pruning inevitably leads to visual information loss by discarding tokens prior to or within early layers, thereby substantially degrading model performance.

\begin{figure*}[ht!]
    \centering
    \setlength{\abovecaptionskip}{0cm}
    \setlength{\belowcaptionskip}{-0.2cm}
    \includegraphics[width=\textwidth]{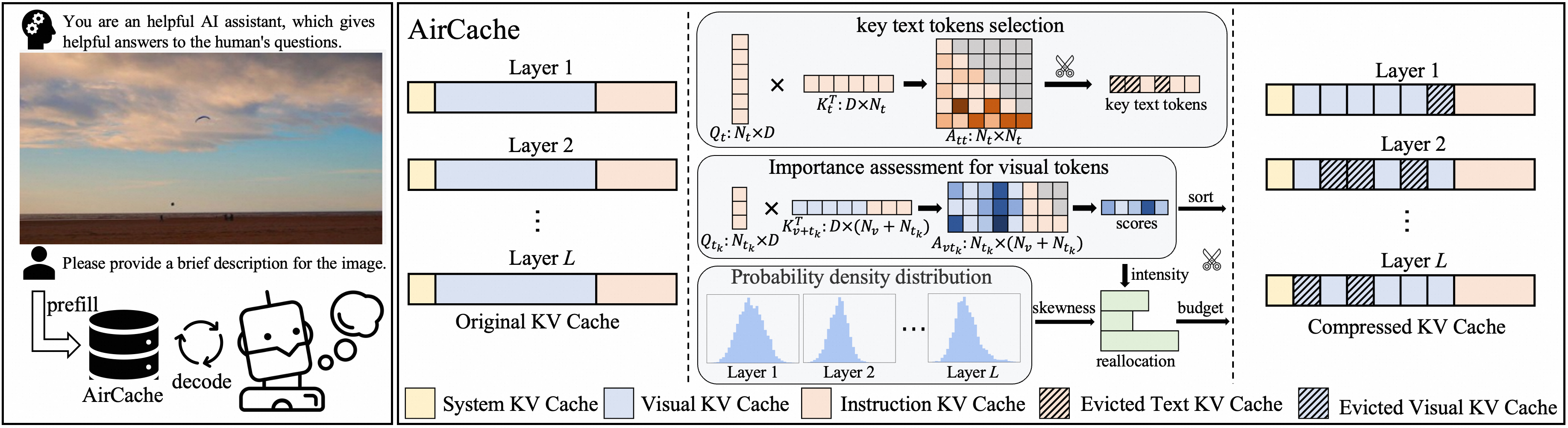}
    \caption{The left panel illustrates the inference workflow of standard LVLMs. The proposed method integrates during the KV cache storage process following the prefill stage, maintaining compatibility with mainstream LVLMs architecture. The right panel presents the overview of AirCache. We first employ self-attention across the instruction text tokens to identify the key text tokens, establishing an elite observation window for assessing visual token significance. We further reallocate the compression budgets between layers based on the strength and skewness of the importance distribution. Finally, pruning is applied to the KV cache according to the importance ranking of visual tokens in the KV Cache and the compression budget to obtain the final compressed KV cache.}
    \vspace{-10pt}
 	\label{fig: overview}
\end{figure*}

{\bf KV cache compression}. KV cache strategies accelerate inference by storing the key and value states of previous tokens, thereby avoiding redundant computations during decoding. However, as the number of tokens increases, the pressure on memory and bandwidth also rises, limiting the model's application in long content understanding or generation. To address this challenge, mainstream methods primarily focus on two directions: intra-layer KV cache pruning \cite{kim2023compressed, nawrot2024dynamic, zhang2024cam} and inter-layer compression budget allocation \cite{liu2025minicache, yang2024kvsharer}. H2O \cite{zhang2024h2o} and SnapKV \cite{li2024snapkv} propose evicting tokens in the KV cache using global and local cumulative attention scores, while PyramidKV \cite{cai2024pyramidkv} constructs a pyramid-like hierarchical budget allocation to achieve more refined KV cache compression management. The input to LVLMs is a cross-modal construct primarily composed of visual tokens with auxiliary text tokens, necessitating additional consideration of the impact of different modality differences when extending KV cache compression to LVLMs. Elastic \cite{liu2024efficient} merges non-critical tokens with important tokens, PrefixKV \cite{wang2024prefixkv} introduces an adaptive layer-wise KV retention recipe based on binary search for maximal preservation of contextual information. VL-Cache \cite{tu2024vl} utilizes visual token sparsity for layer-wise budget allocation and token eviction through dedicated scoring policies. 

\section{Methodology}
\label{sec:method}
\subsection{Preliminary}
Similar to most LLMs, the inference process of LVLMs is primarily divided into prompt prefill and token decoding.

\noindent \textbf{Prompt Prefill.} Given the hidden states $X \in \mathbb{R}^{N \times D}$ of the input prompt, where $N$ is the total number of visual and text prompt tokens, and $D$ denotes hidden dimension. We simplify the expression by omitting the indices of the head and layer of the hidden states. The query, key, and value states of the attention block can be represented as: 
\begin{equation}\label{Eq.1}
    \mathbf{Q}=\mathbf{X} \mathbf{W}_Q, \quad \mathbf{K}=\mathbf{X} \mathbf{W}_K, \quad \mathbf{V}=\mathbf{X} \mathbf{W}_V ,
\end{equation}
where $\mathbf{W}_Q, \mathbf{W}_K, \mathbf{W}_V \in \mathbb{R}^{D \times D}$ are learnable projection matrices. Following the completion of the attention interaction, the key and value are stored in the KV cache to support subsequent token generation.

\noindent \textbf{Token Decoding.} In this stage, the model efficiently generates new tokens by iteratively utilizing and updating the KV cache. Specifically, during the \emph{i}th iteration, we only need to compute the key and value associated with the newly generated token $\mathbf{x}_i \in \mathbb{R}^{1 \times D}$. By employing cache indexing, the model avoids recomputing the key and value for tokens that have already been processed. Subsequently, the KV cache is updated with the newly computed key and value for the latest token:
\begin{equation}\label{Eq.2}
    \scalebox{0.96}{
    $\mathbf{K}=\operatorname{Concat}\left(\mathbf{K}, \mathbf{x}_i \mathbf{W}_K\right), \mathbf{V}=\operatorname{Concat}\left(\mathbf{V}, \mathbf{x}_i \mathbf{W}_V\right).$}
\end{equation}
Although the KV cache mitigates the issue of redundant computation during the decoding process, its characteristic of linear growth in computational requirements relative to the length of the input sequence poses challenges regarding latency and memory usage, especially in scenarios involving longer inputs or outputs. This issue is further exacerbated as recent LVLMs tend to prioritize high-resolution, high-frame-rate visual inputs.

\subsection{Overview}
The AirCache is primarily divided into two components. The first component evaluates the significance of visual tokens based on the elite observation window. Section \ref{sec:elite_observation_window} details how to optimize the relevancy between visual and language modalities to guide the compression of the visual segment in the KV cache. The second component quantifies the KV cache budget requirements across various layers by analyzing the overall strength and skewness of the importance score distribution, thereby facilitating dynamic layer-wise KV cache budget allocation, as elaborated in Section \ref{sec:budget_allocation}. The architecture and algorithm underlying the method are illustrated in Figure \ref{fig: overview} and Algorithm \ref{alg: AirCache}.

\subsection{Elite Observation Window}
\label{sec:elite_observation_window}
\begin{figure*}[t]
    \centering
    \includegraphics[width=0.85\linewidth]{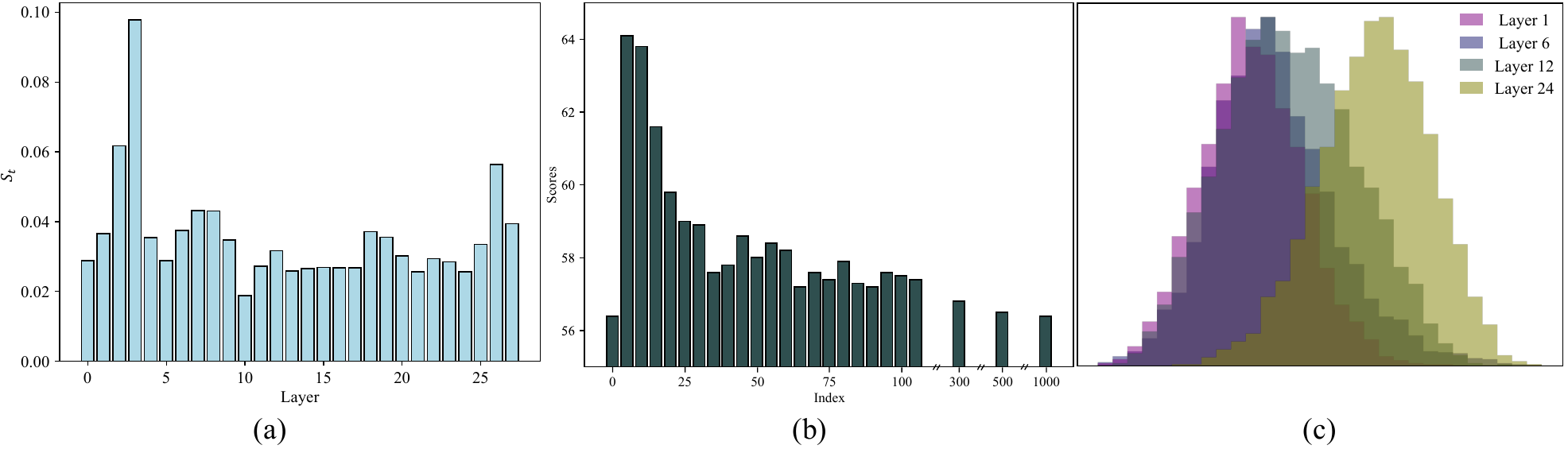}
    \caption{(a) The strength of the distribution of visual token importance scores across different layers, which is defined as the sum of all visual token importance scores. (b) Performance with selecting only one visual token each time based on the sorted results of visual token importance scores, the index of 0 indicates the setting of removing all visual tokens. (c) The distribution of visual token importance scores across different layers, with only selected layers displayed for clarity. All results are derived from LLaVA-OV-7B \cite{li2024llava} on the ChatQA \cite{masry2022chartqa}.}
    \label{fig: overlap}
    \vspace{-10pt}
\end{figure*}

In LVLMs, the presence of multiple modalities complicates the selection of observation windows, suggesting that adhering to the relevant practices established in LLMs may not be optimal. For instance, as illustrated in Figure \ref{fig1_motivation}, the visual token importance measures derived from the continuous last segment of the prompt exhibit poor consistency. This variability indicates that different text tokens within the observation window yield significantly diverse visual tokens importance distributions. Consequently, the hit rate statistics derived from the voting mechanism are prone to substantial noise, and the disparities among modalities may further amplify this noise interference. Based on this consideration, we propose refining the observation windows by incorporating carefully selected text tokens. By utilizing the attention hit rate within the text modality, we aim to enhance the consistency of visual token evaluation, thereby improving the performance of KV cache compression. To this end, we reorganize the hidden states of the input prompt by:

\begin{equation}\label{Eq.121}
    \mathbf{X}=\operatorname{Concat}(\mathbf{X}_v,\mathbf{X}_t) \in \mathbb{R}^{(N_v+N_t) \times D},
\end{equation}
where $\mathbf{X}_v$ and $\mathbf{X}_t$ represent the hidden states corresponding to visual tokens and instruction text tokens, respectively. $N_v$ and $N_t$ denote the respective quantities of these tokens, satisfying the condition $N_v+N_t=N$. For convenience, the system prompts are omitted. We compute the text unimodal attention matrix composed of instruction text tokens as:
\begin{equation}\label{Eq.121}
    \mathbf{A}_{tt}=\operatorname{Softmax}\left(\frac{\mathbf{Q}_t \mathbf{K}_t^T}{\sqrt{D}}\right) \in \mathbb{R}^{N_t \times N_t},
\end{equation}
where $\mathbf{Q}_t$ and $\mathbf{K}_t$ denote the query and key of the text tokens, respectively. Considering the characteristics of causal attention, we select the last token of the instruction text as reference and filter out key text tokens that receive high attention scores based on their interactions with other text tokens. Consistent with \cite{zhang2024h2o}, we employ a relevance threshold to regulate the sparsity of key text tokens:
\begin{equation}\label{Eq.05}
     k =\left\{j \mid \mathbf{A}_{tt}[N_t-1, j] \geq \alpha \cdot \max \mathbf{A}_{tt}\left[N_t-1, :\right]\right\},
\end{equation}
where $\alpha \in [0,1]$ represents the relevance threshold, and $j \in \{0,1,\ldots, N_t-1\}$ is the index of the text token. When $\alpha$ is equal to 0, all text tokens are retained, and when $\alpha$ is equal to 1, only the text token with the highest score is preserved. We redefine the key for the aggregated visual tokens and key text tokens as $\mathbf{K}_{vt_k}$, along with the query corresponding to the key text tokens $\mathbf{Q}_{t_k}$, as follows:
\begin{equation}\label{Eq.121}
\begin{array}{c}
     \quad \mathbf{Q}_{t_k}=\mathbf{Q}_t[k]\in \mathbb{R}^{N_{t_{k}} \times D}, \\
     \mathbf{K}_{vt_k}=\operatorname{Concat}(\mathbf{K}_v, \mathbf{K}_t[k])\in \mathbb{R}^{(N_v + N_{t_{k}}) \times D}, 
\end{array}
\end{equation}
where $\mathbf{K}_v$ represents the key of the visual token, and $N_{t_{k}}$ denotes the number of selected key text tokens. Furthermore, the attention matrix between the visual tokens and the key text tokens can be computed as follows:
\begin{equation}\label{Eq.121}
\mathbf{A}_{vt_{k}}=\operatorname{Softmax}\left(\frac{\mathbf{Q}_{t_k}\mathbf{K}_{vt_k}^T}{\sqrt{D}}\right) \in \mathbb{R}^{N_{t_k} \times (N_v+N_{t_{k}})}.
\end{equation}
The elite observation window, comprising only these key text tokens, offers a more comprehensive and focused semantic representation compared to those formed by all text tokens or fixed continuous sequences. It provides enhanced stability and accuracy in assessing the importance of visual tokens. The final importance scores of the visual tokens can be obtained by performing average pooling of the attention matrix along the text dimension:
\begin{equation}\label{Eq.121}
     \mathbf{I}_v =\frac{1}{N_{t_k}} \sum_{j=0}^{N_{t_k}-1} \mathbf{A}_{vt_k}[j,:N_v].
\end{equation}
By sorting the importance scores, the KV cache of visual tokens with lower rankings can be discarded in accordance with the compression budget. Furthermore, the elite observation window we employ exhibits reduced computational complexity, and the additional overhead of key text token selection is negligible in visual token-heavy LVLMs.

\subsection{Layer-wise KV Cache Budget Allocation}
\label{sec:budget_allocation}
We further investigate the characteristics of the visual token importance score distribution based on the elite observation window across various layers. Specifically, we consider the sum of the importance scores of all visual tokens as an indicator of the layer's attention to visual information. As illustrated in Figure \ref{fig: overlap}(a), there is a significant disparity in the attention given to visual tokens across different layers. This observation highlights the necessity of performing layer-wise token budget allocation, where layers dedicating more attention to visual information merit a larger token budget. As shown in Figure \ref{fig: overlap}(b), we select one visual token at a time based on the importance scores. It can be observed that only approximately 10\% of all visual tokens have a significantly positive impact on the final result. This suggests a head effect in the model's attention distribution over visual tokens, with a small subset receiving higher attention, while the majority have lower and more balanced scores. Extending to the importance distribution across different layers in Figure \ref{fig: overlap}(c), it shows significant differences in importance distribution among the layers, particularly regarding the head effect of visual tokens. Consequently, we aim to allocate a larger token budget to the layers that demonstrate a pronounced head effect in their visual tokens importance distribution. This indicates a more accurate and nuanced understanding of the visual content by those layers.

As aforementioned, we quantify the allocation of visual token budget across different layers from two perspectives: the strength and skewness of the importance distribution. The strength of the importance distribution is calculated as:
\begin{equation}\label{Eq.xxx}
    s_t =\sum_{i=0}^{N_v-1} \mathbf{I}_v[i].
\end{equation}
We exclude the attention between text tokens and instead aggregate the attention scores between text tokens and all visual tokens to quantify the importance of visual information in each layer. The greater the importance, the more budget is allocated. Meanwhile, we use the skewness of importance distribution to assess how effectively each layer aligns with the distribution characteristics of attention to visual tokens:
\begin{equation}\label{Eq.xxx}
    s_k=\frac{N_v}{(N_v-1)(N_v-2)} \sum_{i=1}^{N_{v}}\left(\frac{\mathbf{I}_v[i]-\mu_{I_v}}{\sigma_{I_v}}\right)^3,
\end{equation}
where $\mu_{I_v}$ and $\sigma_{I_v}$ represent the mean and standard deviation of the importance as well as attention scores distribution of visual tokens, which essentially reflect the distribution of attention scores. Ultimately, the reallocated token budget integrates these factors into the original budget $r$:

\begin{algorithm}[!ht]
    \caption{AirCache for KV Cache Compression. 
    }
    \label{alg: AirCache}
    \begin{algorithmic}[1]
        \small
        \State \textbf{Input:} Total cache budget $r$, key, query, value of text prompt tokens $K_t$, $Q_t$, $V_t \in \mathbb{R}^{N_{t} \times D}$, key, query, value of visual prompt tokens $K_v$, $Q_v$, $V_v \in \mathbb{R}^{N_{v} \times D}$, number of layers L. 
        \For{layer $l \gets 0$ to $N-1$}
            \State $A_{tt} \leftarrow \operatorname{Softmax}\left(\frac{\mathbf{Q}_t \mathbf{K}_t^T}{\sqrt{D}}\right)$
            \State Obtain the index of key text tokens $k$ using Eq. (5).
            \State $Q_{t_k}, K_{vt_k} \leftarrow Q_t[k], \operatorname{Concat}(K_v, K_t[k])$
            \State $\mathbf{A}_{vt_{k}} \leftarrow \operatorname{Softmax}\left(\frac{\mathbf{Q}_{t_k}\mathbf{K}_{vt_k}^T}{\sqrt{D}}\right)$
            \State $\mathbf{I}_v \leftarrow \frac{1}{N_{t_k}} \sum_{j=0}^{N_{t_k}-1} \mathbf{A}_{vt_k}[j,:N_v]$ \Comment{Importance Scores}
            \State $s_t \leftarrow \sum_{i=0}^{N_v-1} \mathbf{I}_v[i]$ \Comment{Strength}
            \State $s_k \leftarrow \frac{N_v}{(N_v-1)(N_v-2)} \sum_{i=1}^{N_{v}}\left(\frac{\mathbf{I}_v[i]-\mu_{I_v}}{\sigma_{I_v}}\right)^3$ \Comment{Skewness}
            \State $C^l \leftarrow (\{(K_v, K_t), (V_v, V_t), (\mathbf{I}_v, s_t, s_k)\})$ 
        \EndFor
        
        \For{layer $l \gets 0$ to $N-1$}
            \State $K_v, K_t, V_v, V_t, \mathbf{I}_v, s_t, s_k \leftarrow C^l$
            \State $s_{t}^{'}, s_{k}^{'} \leftarrow s_t, s_k$ \Comment{Normalized Across Different Layers}
            \State $\hat{r}=\frac{1}{2}(s^{'}_t+s^{'}_k)r$ \Comment{Obtain Layer-wise Budget}
            \State $K^{'} \leftarrow \operatorname{Concat}(K_v(\mathbf{I}[N_v\times\hat{r}]), K_t)$ \Comment{Compress Key}
            \State $V^{'} \leftarrow \operatorname{Concat}(V_v(\mathbf{I}[V_v\times\hat{r}]), V_t)$ \Comment{Compress Value}
            \State $C^l \leftarrow (\{K^{'}, V^{'}\})$
        \EndFor
        \State \textbf{Return:} $C$ \Comment{Compressed KV Cache}
    \end{algorithmic}
\end{algorithm}
\vspace{-10pt}

\begin{equation}\label{Eq.xxx}
    \hat{r}=\frac{1}{2}(s^{'}_t+s^{'}_k)r.
\end{equation}
where $s^{'}_t$ and $s^{'}_k$ represent the normalized strength and skewness across different layers, respectively.

\section{Experiments}
\label{sec:ex}
\subsection{Setup}
\label{sec:ID}
\noindent {\bf Models.} The selection of LVLMs is primarily based on two criteria: model architecture and parameter size. In terms of architecture, the objective is to validate the effectiveness of the proposed method across various architectures, including LLaVA-OV-7B \cite{li2024llava}, InternVL2-8B \cite{chen2024internvl}, and Qwen2-VL-7B \cite{wang2024qwen2}. These models exhibit significant differences in visual token encoding, ViT, and LLM. For example, both LLaVA-OV-7B \cite{li2024llava} and InternVL2-8B \cite{chen2024internvl} utilize the AnyRes strategy with distinct ViT to increase the number of visual tokens, while Qwen2-VL-7B \cite{wang2024qwen2} supports native resolution with a large quantity of visual tokens. In terms of parameter size, models of InternVL2 series are selected for their diverse range of parameter scales, including InternVL2-1B \cite{chen2024internvl}, InternVL2-4B \cite{chen2024internvl}, and InternVL2-26B \cite{chen2024internvl}. Due to space constraints, results for this section are provided in the appendix. Additionally, LLaVA-Video-7B-Qwen2 \cite{zhang2024video} serves as the base model for the video tasks evaluation.

\begin{table*}[th]
\caption{The comparison of the KV cache compression methods on multimodal VQA benchmarks. The best result is highlighted in bold.}
\resizebox{1.0\linewidth}{!}{
\begin{tabular}{c|c|cccc|cccc|cccc|cccc}
\hline
    & & \multicolumn{4}{c|}{ChatQA \cite{masry2022chartqa}} & \multicolumn{4}{c|}{InfoVQA \cite{mathew2022infographicvqa}} & \multicolumn{4}{c|}{DocVQA \cite{mathew2021docvqa}} & \multicolumn{4}{c}{TextVQA \cite{singh2019towards}} \\ \cline{3-18} 
    \multirow{-2}{*}{Models} & \multirow{-2}{*}{Methods} & 50\% & 10\% & 5\% & 1\% & 50\% & 10\% & 5\% & 1\% & 50\% & 10\% & 5\% & 1\% & 50\% & 10\% & 5\% & 1\% \\ \hline
    & \cellcolor[HTML]{EFEFEF}Full & \cellcolor[HTML]{EFEFEF}80.3 & \cellcolor[HTML]{EFEFEF}80.3 & \cellcolor[HTML]{EFEFEF}80.3 & \cellcolor[HTML]{EFEFEF}80.3 & \cellcolor[HTML]{EFEFEF}66.1 & \cellcolor[HTML]{EFEFEF}66.1 & \cellcolor[HTML]{EFEFEF}66.1 & \cellcolor[HTML]{EFEFEF}66.1 & \cellcolor[HTML]{EFEFEF}87.0 & \cellcolor[HTML]{EFEFEF}87.0 & \cellcolor[HTML]{EFEFEF}87.0 & \cellcolor[HTML]{EFEFEF}87.0 & \cellcolor[HTML]{EFEFEF}76.0 & \cellcolor[HTML]{EFEFEF}76.0 & \cellcolor[HTML]{EFEFEF}76.0 & \cellcolor[HTML]{EFEFEF}76.0 \\
     & H2O \cite{zhang2024h2o} & 79.8 & 77.4 & 74.0 & 71.0 & 65.4 & 59.2 & 54.2 & 52.0 & 86.4 & 74.2 & 60.8 & 55.3 & 75.5 & 70.1 & 67.9 & 60.4 \\
     & Elastic \cite{liu2024efficient} & 80.0 & 77.9 & 76.5 & 71.0 & 65.3 & 60.5 & 57.2 & 52.0 & 86.6 & 74.2 & 60.8 & 55.3 & 75.5 & 72.0 & 68.7 & 60.2 \\
     & PrefixKV \cite{wang2024prefixkv} & 80.1 & 78.2 & 76.8 & 70.9 & 65.4 & 61.1 & 58.0 & 51.9 & 86.6 & 80.5 & 74.7 & 55.4 & 75.8 & 72.7 & 70.0 & 60.2 \\
     & SnapKV \cite{li2024snapkv} & 80.1 & 79.3 & 78.3 & 72.9 & \bf 66.1 & 64.2 & 63.0 & 57.8 & \bf 86.9 & 84.4 & 80.6 & 64.1 & 75.8 & 73.4 & 69.5 & 58.2 \\
    \multirow{-6}{*}{LLaVA-OV-7B \cite{li2024llava}}  & Ours & \bf 80.4 & \bf 79.9 & \bf 79.4 & \bf 76.4 & 66.0 & \bf 65.7 & \bf 64.6 & \bf 62.5 & 86.8 & \bf 85.5 & \bf 83.3 & \bf 73.2 & \bf 76.0 & \bf 75.3 & \bf 73.4 & \bf 67.1 \\ \hline \hline 
    & \cellcolor[HTML]{EFEFEF}Full & \cellcolor[HTML]{EFEFEF}82.2 & \cellcolor[HTML]{EFEFEF}82.2 & \cellcolor[HTML]{EFEFEF}82.2 & \cellcolor[HTML]{EFEFEF}82.2 & \cellcolor[HTML]{EFEFEF}73.0 & \cellcolor[HTML]{EFEFEF}73.0 & \cellcolor[HTML]{EFEFEF}73.0 & \cellcolor[HTML]{EFEFEF}73.0 & \cellcolor[HTML]{EFEFEF}91.0 & \cellcolor[HTML]{EFEFEF}91.0 & \cellcolor[HTML]{EFEFEF}91.0 & \cellcolor[HTML]{EFEFEF}91.0 & \cellcolor[HTML]{EFEFEF}77.7 & \cellcolor[HTML]{EFEFEF}77.7 & \cellcolor[HTML]{EFEFEF}77.7 & \cellcolor[HTML]{EFEFEF}77.7 \\
     & H2O \cite{zhang2024h2o} & 81.9 & 81.6 & 79.3 & 76.8 & 72.3 & 70.6 & 65.1 & 61.2 & 90.5 & 79.4 & 70.3 & 48.9 & 77.1 & 70.0 & 66.0 & 59.6 \\
     & Elastic \cite{liu2024efficient} & 81.8 & 81.0 & 79.4 & 76.8 & 72.3 & 67.0 & 65.1 & 60.8 & 90.5 & 79.4 & 70.3 & 53.5 & 77.0 & 70.2 & 66.0 & 59.0\\
     & PrefixKV \cite{wang2024prefixkv} & 81.9 & 80.7 & 79.3 & 76.7 & 72.3 & 67.4 & 65.3 & 61.0 & 90.6 & 79.5 & 70.2 & 55.1 & 77.1 & 70.3 & 66.2 & 58.6\\
     & SnapKV \cite{li2024snapkv} & 82.2 & 80.4 & 79.5 & 75.0 & 72.8 & 72.3 & 68.4 & 64.9 & 90.9 & \bf 90.1 & 83.5 & 71.1 & 77.4 & 73.9 & 71.0 & 61.7 \\
    \multirow{-6}{*}{InternVL2-8B \cite{chen2024internvl}} & Ours & \bf 82.2 & \bf 81.7 & \bf 81.0 & \bf 78.4 & \bf 73.2 & \bf 72.6 & \bf 71.8 & \bf 68.2 & \bf 91.0 & 90.0 & \bf 88.5 & \bf 78.5 & \bf 77.5 & \bf 77.0 & \bf 75.4 & \bf 68.6 \\ \hline \hline
     & \cellcolor[HTML]{EFEFEF}Full & \cellcolor[HTML]{EFEFEF}83.2 & \cellcolor[HTML]{EFEFEF}83.2 & \cellcolor[HTML]{EFEFEF}83.2 & \cellcolor[HTML]{EFEFEF}83.2 & \cellcolor[HTML]{EFEFEF}76.2 & \cellcolor[HTML]{EFEFEF}76.2 & \cellcolor[HTML]{EFEFEF}76.2 & \cellcolor[HTML]{EFEFEF}76.2 & \cellcolor[HTML]{EFEFEF}93.8 & \cellcolor[HTML]{EFEFEF}93.8 & \cellcolor[HTML]{EFEFEF}93.8 & \multicolumn{1}{l|}{\cellcolor[HTML]{EFEFEF}93.8} & \multicolumn{1}{l}{\cellcolor[HTML]{EFEFEF}84.4} & \multicolumn{1}{l}{\cellcolor[HTML]{EFEFEF}84.4} & \multicolumn{1}{l}{\cellcolor[HTML]{EFEFEF}84.4} & \multicolumn{1}{l}{\cellcolor[HTML]{EFEFEF}84.4} \\
     & H2O \cite{zhang2024h2o} & 83.2 & 78.9 & 75.0 & 66.4 & 75.4 & 65.7 & 61.0 & 55.1 & 90.5 & 82.3 & 70.7 & 48.9 & 84.2 & 75.6 & 68.2 & 54.5 \\
     & Elastic \cite{liu2024efficient} & 83.2 & 79.7 & 75.4 & 68.0 & 75.5 & 67.1 & 62.3 & 55.5 & 93.5 & 83.0 & 71.9 & 49.2 & 84.2 & 76.7 & 68.8 &54.4  \\
     & PrefixKV \cite{wang2024prefixkv} & 83.2 & 79.4 & 75.9 & 68.0 & 75.7 & 67.9 & 62.7 & 55.7 & 93.7 & 83.7 & 72.8 & 49.8 & 84.2 & 76.9 & 69.4 & 54.3 \\
     & SnapKV \cite{li2024snapkv} & 83.3 & 81.6 & 80.0 & 69.1 & \bf 76.3 & 74.9 & 73.0 & 66.2 & 93.7 & 87.2 & 87.1 & 71.2 & 84.3 & 83.1 & 74.6 & 60.2 \\
    \multirow{-6}{*}{Qwen2-VL-7B \cite{wang2024qwen2}} & Ours & \bf 83.4 & \bf 82.3 & \bf 80.8 & \bf 75.9 & 76.2 & \bf 75.2 & \bf 74.5 & \bf 70.7 & \bf 93.8 & \bf 92.9 & \bf 90.9 & \bf 78.3 & \bf 84.4 & \bf 83.4 & \bf 80.9 & \bf 68.6 \\ \hline
\end{tabular}
}
\label{tab: main}
\vspace{-10pt}
\end{table*}

\begin{table}[]
\caption{Performance comparison on other types datasets. Best-performing results are highlighted in bold.}
\resizebox{1.0\linewidth}{!}{
\begin{tabular}{cccccccccc}
\hline
\multirow{2}{*}{Methods} & \multicolumn{3}{c}{MMBench-EN \cite{liu2024mmbench}} & \multicolumn{3}{c}{MME \cite{fu2024mme}} & \multicolumn{3}{c}{MMBench-Video \cite{fang2024mmbench}} \\ \cmidrule(l){2-4} \cmidrule(l){5-7} \cmidrule(l){8-10}
 & 50\% & 10\% & 1\% & 50\% & 10\% & 1\% & 50\% & 10\% & 1\% \\ \hline
\cellcolor[HTML]{EFEFEF} Full & \cellcolor[HTML]{EFEFEF} 82.3 & \cellcolor[HTML]{EFEFEF} 82.3 & \cellcolor[HTML]{EFEFEF} 82.3 & \cellcolor[HTML]{EFEFEF} 1585 & \cellcolor[HTML]{EFEFEF} 1585 & \cellcolor[HTML]{EFEFEF} 1585 & \cellcolor[HTML]{EFEFEF} 1.81 & \cellcolor[HTML]{EFEFEF} 1.81 & \cellcolor[HTML]{EFEFEF} 1.81 \\
H2O \cite{zhang2024h2o}& 0.82 & 0.82 & 0.82 & 1585 & 1585 & 1585 & 1.66  & 1.63  & 1.47\\
Elastic \cite{liu2024efficient} & 0.82 & 0.82 & 0.82 & 1585 & 1585 & 1585 & 1.68 & 1.62 & 1.51\\
PrefixKV \cite{wang2024prefixkv} & 0.82 & 0.82 & 0.82 & 1585 & 1585 & 1585 & 1.72 & 1.68 & 1.50 \\
SnapKV \cite{li2024snapkv} & 0.82 & 0.82 & 0.82 & 1585 & 1585 & 1585 & 1.75 & 1.70 & 1.55\\
Ours & 0.82 & 0.82 & 0.82 & 1585 & 1585 & 1585 & 1.80 & 1.78 & 1.67\\ \hline
\end{tabular}
}
\vspace{-10pt}
\label{tab: other_type_datasets}
\end{table}

\noindent {\bf Datasets.} The performance is primarily evaluated across four types of datasets: 1) Visual Question Answering (VQA) datasets, which include ChatQA \cite{masry2022chartqa}, InfoVQA \cite{mathew2022infographicvqa}, DocVQA \cite{mathew2021docvqa}, and TextVQA \cite{singh2019towards}. Since KV cache compression mainly affects the decoding stage, VQA datasets that require longer output tokens are preferred for evaluating the method's efficacy. The similarity between generated and reference answers is typically measured by the Average Normalized Levenshtein Similarity (ANLS) metric \cite{biten2019scene}. 2) Multiple-choice question datasets, such as MMBench-EN \cite{liu2024mmbench}, involves selecting the correct option from several choices, with accuracy serving as the ultimate metric. 3) Judgment datasets like MME \cite{fu2024mme}, also utilize judgment accuracy as the evaluation criterion. 4) Video VQA datasets, such as MMBench-Video \cite{fang2024mmbench}, measure the alignment between generated and reference answers using metrics provided by a large model evaluator.

\noindent {\bf Baselines.} The comparison focuses on two main categories of methods. The first category encompasses classic and prominent KV cache compression techniques in LLMs, such as H2O \cite{zhang2024h2o}, and SnapKV \cite{li2024snapkv}. The second category comprises methods specifically optimized for LVLMs, namely PrefixKV \cite{wang2024prefixkv} and Elastic Cache \cite{liu2024efficient}. H2O \cite{zhang2024h2o} employs a full-range approach, while SnapKV \cite{li2024snapkv} adopts a window-based strategy using textual tokens as the observation window for evaluating token importance. PrefixKV \cite{wang2024prefixkv} utilizes the Lorenz curve to characterize the importance distribution across different layers, guided by the retained important KV matrix. Elastic Cache \cite{liu2024efficient} employs a fixed-point elimination strategy for the token importance with cache merging. All KV cache compression methods are applied exclusively to the visual tokens, as experiments have demonstrated that it offers greater stability and better performance compared to compressing the entire tokens.

\noindent {\bf Implementation details.} All evaluation results are derived from VLMEvalKit \cite{duan2024vlmevalkit}, which is an open-source evaluation toolkit for LVLMs. The relevance threshold in Equation \ref{Eq.05} is set to 0.9. The referee model for MMBench-Video \cite{fang2024mmbench} is Qwen2.5-72B \cite{yang2024qwen2}. The proposed method performs a one-time KV cache compression only after the prefill stage is completed. Specifically, once the prefill ends for each layer, the complete KV cache is fully saved. After computing all layers, we then calculate the budget for each layer and compress the KV cache accordingly. All experiments are conducted on 8$\times$A100-80G GPUs.

\subsection{Main Results}
We report the results in four sections: VQA datasets, other types of datasets including multiple-choice questions, judgment, and video VQA, inference performance, and the comparison with token pruning methods.

\begin{table*}[]
\centering
\caption{Quantitative results on inference latency and throughput. The number of tokens output is consistently set to 512.}
\resizebox{0.9\linewidth}{!}{
\begin{tabular}{cccccccccc}
\hline
& & \multicolumn{2}{c}{Prefill Latency (s)} & \multicolumn{3}{c}{Decoding Latency (s)} & \multicolumn{3}{c}{Throughput (token/s)} \\ \cmidrule(l){3-4} \cmidrule(l){5-7} \cmidrule(l){8-10}
\multirow{-2}{*}{Batch Size} & \multirow{-2}{*}{Prompt Length} & \multicolumn{1}{c}{Full} & \multicolumn{1}{c}{Ours} & \multicolumn{1}{c}{Full} & \multicolumn{1}{c}{50\%} & 10\% & \multicolumn{1}{c}{Full} & \multicolumn{1}{c}{50\%} & \multicolumn{1}{c}{10\%} \\ \hline
& 2k & \cellcolor[HTML]{EFEFEF} 1.6 & 1.8\(_{\color{blue}{\tiny+12.5\%}}\) & \cellcolor[HTML]{EFEFEF} 11.6 & 9.4\(_{\color{blue}{\tiny-19.0\%}}\) & 8.2\(_{\color{blue}{\tiny-29.3\%}}\) & \cellcolor[HTML]{EFEFEF} 353 & 436\(_{\color{blue}{\tiny+23.5\%}}\) & 500\(_{\color{blue}{\tiny+41.6\%}}\) \\
& 8k & \cellcolor[HTML]{EFEFEF} 4.9 & 5.3\(_{\color{blue}{\tiny+8.2\%}}\) & \cellcolor[HTML]{EFEFEF} 15.0 & 11.9\(_{\color{blue}{\tiny-20.7\%}}\) & 9.4\(_{\color{blue}{\tiny-37.3\%}}\) & \cellcolor[HTML]{EFEFEF} 273 & 344\(_{\color{blue}{\tiny+26.0\%}}\) & 436\(_{\color{blue}{\tiny+59.7\%}}\) \\
& 16k & \cellcolor[HTML]{EFEFEF} 11.3 & 11.9\(_{\color{blue}{\tiny+5.3\%}}\) & \cellcolor[HTML]{EFEFEF} 21.7 & 15.0\(_{\color{blue}{\tiny-30.9\%}}\) & 10.4\(_{\color{blue}{\tiny-47.5\%}}\) & \cellcolor[HTML]{EFEFEF} 189 & 273\(_{\color{blue}{\tiny+44.4\%}}\) & 394\(_{\color{blue}{\tiny+108.5\%}}\) \\
\multirow{-4}{*}{8} 
& 32k & \cellcolor[HTML]{EFEFEF} 25.2 & 26.4\(_{\color{blue}{\tiny+4.8\%}}\) & \cellcolor[HTML]{EFEFEF} 35.9 & 22.0\(_{\color{blue}{\tiny-38.7\%}}\) & 12.3\(_{\color{blue}{\tiny-65.7\%}}\) & \cellcolor[HTML]{EFEFEF} 114 & 186\(_{\color{blue}{\tiny+63.2\%}}\) & 333\(_{\color{blue}{\tiny+192.1\%}}\)\\ \hline
& 2k & \cellcolor[HTML]{EFEFEF} 2.3 & 2.5\(_{\color{blue}{\tiny+8.7\%}}\) & \cellcolor[HTML]{EFEFEF} 14.4 & 11.2\(_{\color{blue}{\tiny-22.2\%}}\) & 9.7\(_{\color{blue}{\tiny-32.6\%}}\) & \cellcolor[HTML]{EFEFEF} 569 & 731\(_{\color{blue}{\tiny+28.5\%}}\) & 845\(_{\color{blue}{\tiny+48.5\%}}\) \\
& 8k & \cellcolor[HTML]{EFEFEF} 9.4 & 9.8\(_{\color{blue}{\tiny+4.3\%}}\) & \cellcolor[HTML]{EFEFEF} 22.6 & 15.6\(_{\color{blue}{\tiny-31.0\%}}\) & 11.8\(_{\color{blue}{\tiny-47.8\%}}\) & \cellcolor[HTML]{EFEFEF} 362 & 525\(_{\color{blue}{\tiny+45.0\%}}\) & 694\(_{\color{blue}{\tiny+91.7\%}}\)\\
\multirow{-3}{*}{16} 
& 16k & \cellcolor[HTML]{EFEFEF} 20.3 & 21.4\(_{\color{blue}{\tiny+5.4\%}}\) & \cellcolor[HTML]{EFEFEF} 36.9 & 23.0\(_{\color{blue}{\tiny-37.7\%}}\) & 12.8\(_{\color{blue}{\tiny-65.3\%}}\) & \cellcolor[HTML]{EFEFEF} 222 & 356\(_{\color{blue}{\tiny+60.4\%}}\) & 640\(_{\color{blue}{\tiny+188.3\%}}\) \\ \hline
\end{tabular}
}
\label{tab: latency_throughput}
\end{table*}

\begin{table*}[]
\centering
\caption{Performance comparison with token pruning methods under various compression ratio. The prefill and decoding latency are measured under the settings of batch size of 16, total input token of 8k, and the compression ratio of 10\%.}
\resizebox{1.0\linewidth}{!}{
\begin{tabular}{ccccccccccccccc}
\hline
\multirow{2}{*}{Methods} & \multicolumn{3}{c}{ChatQA \cite{masry2022chartqa}} & \multicolumn{3}{c}{MMBench-EN \cite{liu2024mmbench}} & \multicolumn{3}{c}{MME \cite{fu2024mme}} & \multicolumn{3}{c}{MMBench-Video \cite{fang2024mmbench}} & \multicolumn{2}{c}{Latency (s)} \\ \cmidrule(l){2-4} \cmidrule(l){5-7} \cmidrule(l){8-10} \cmidrule(l){11-13} \cmidrule(l){14-15}
& 50\%    & 10\%    & 1\%    & 50\%      & 10\%     & 1\%     & 50\%   & 10\%   & 1\%   & 50\%       & 10\%       & 1\%      & Prefill        & Decoding       \\ \hline
\cellcolor[HTML]{EFEFEF} Full & \cellcolor[HTML]{EFEFEF} 80.3 & \cellcolor[HTML]{EFEFEF} 80.3 & \cellcolor[HTML]{EFEFEF} 80.3 & \cellcolor[HTML]{EFEFEF} 82.3 & \cellcolor[HTML]{EFEFEF} 82.3 & \cellcolor[HTML]{EFEFEF} 82.3 & \cellcolor[HTML]{EFEFEF} 1585 & \cellcolor[HTML]{EFEFEF} 1585 & \cellcolor[HTML]{EFEFEF} 1585 & \cellcolor[HTML]{EFEFEF} 1.81 & \cellcolor[HTML]{EFEFEF} 1.81 & \cellcolor[HTML]{EFEFEF} 1.81 & \cellcolor[HTML]{EFEFEF} 9.4 & \cellcolor[HTML]{EFEFEF} 22.6 \\
FastV \cite{chen2024image} & 73.6 & 47.7 & 16.9 & 81.4 & 74.8 & 33.6 & 1574 & 1378 & 786 & 1.69 & 1.36 & 1.02 & 5.4 & 11.7 \\
FasterVLM \cite{zhang2024cls} & 75.5 & 50.1 & 18.8 & 81.4 & 77.0 & 26.5 & 1559 & 1441 & 731 & 1.71 & 1.42 & 1.17 & 5.2 & 11.3 \\
IVTP \cite{huang2024ivtp} & 74.9 & 55.8 & 22.5 & 81.7 & 77.5 & 36.2 & 1563 & 1469 & 849 & 1.70 & 1.47 & 1.24 & 5.8 & 12.6 \\
Ours & 80.4 & 79.9 & 76.4 & 82.3 & 82.3 & 82.3 & 1585 & 1585 & 1585 & 1.79 & 1.72 & 1.60 & 9.8 & 11.8 \\ \hline
\end{tabular}
}
\vspace{-10pt}
\label{tab: token_pruning_methods}
\end{table*}

\noindent {\bf VQA datasets.} Table \ref{tab: main} presents the comparative results on the ChatQA \cite{masry2022chartqa}, InfoVQA \cite{mathew2022infographicvqa}, DocVQA \cite{mathew2021docvqa}, and TextVQA \cite{singh2019towards} datasets. These datasets encompass tasks such as conventional image question answering, image information extraction, document image understanding, and text recognition within images. It can be observed that the proposed method achieves the best results under most reduction ratios. When 50\% of visual KV cache is pruned, our method performs almost equivalently to the full KV cache, indicating near lossless performance. Even when reduced to only 10\% of visual KV cache, the performance gap with the full cache is limited to approximately 1\%. As the token compression ratio increases, the advantage of the proposed method becomes more pronounced. For instance, when only 1\% of visual tokens are retained, we outperform SnapKV \cite{li2024snapkv} by an average of 6.5\% across four VQA datasets with the LLaVA-OV-7B \cite{li2024llava}. Moreover, the performance differences across different types of VQA datasets indicate that image documents or text understanding, which require higher local visual perception, have stricter demands on KV cache compression than general image understanding. The proposed method demonstrates a more significant advantage over other methods with these datasets, providing strong evidence of its superiority. Furthermore, the above conclusions also apply to different LVLMs, we achieves a significant advantage with InternVL2-8B \cite{chen2024internvl} and Qwen2-VL-7B \cite{wang2024qwen2}.

\noindent {\bf Other types datasets.} Table \ref{tab: other_type_datasets} presents the results of various methods on other types of multimodal datasets. For datasets involving multiple-choice questions and judgment types, such as MMBench-EN \cite{liu2024mmbench} and MME \cite{fu2024mme}, the final result heavily depends on the accuracy of decoding the first token. Since the KV cache only influences the tokens decoded after the initial token, all methods achieve the same performance as the full KV cache for these types of datasets. The results from the MMBench-Video \cite{fang2024mmbench} dataset indicate that our proposed method achieves superior performance, further demonstrating its generality and superiority.

\noindent {\bf Inference efficiency.} We maintain a constant output token count of 512 and compare the full cache with the proposed AirCache across different batch sizes and input token counts in terms of prefill latency, decoding latency, and throughput. To facilitate testing, the text instruction token count is set to 64. As shown in Table \ref{tab: latency_throughput}, the proposed method significantly improves inference decoding speed under various inputs compared to the full cache. When compressing 50\% of the visual KV cache, the proposed method minimally impacts model performance and prefill latency, while reducing decoding latency by 19\% to 39\% and increasing model throughput by 24\% to 63\%. When compressing 10\% of the visual KV cache, the proposed method slightly decreases model performance by less than 1\%, while further reducing decoding latency by 29\% to 66\% and increasing model throughput by 42\% to 192\%. These results demonstrate that our proposed method can maintain strong model performance while ensuring significant decoding acceleration, indicating the effectiveness of visual token importance assessment and hierarchical budget allocation.

\noindent {\bf Compared with token pruning methods.} In addition to KV cache compression, another important method for accelerating LVLMs inference is to prune visual tokens before the prefill stage. Table \ref{tab: token_pruning_methods} compares the proposed method with these type of acceleration methods, including FastV \cite{chen2024image}, FasterVLM \cite{zhang2024cls}, and IVTP \cite{huang2024ivtp}. A notable characteristic is that the token pruning methods exhibit acceleration effects during the prefill stage as they preemptively prune visual tokens. In contrast, all KV cache compression methods, including the proposed approach, demonstrate prefill stage inference speeds comparable to those of the full cache. However, comparisons across various evaluation sets clearly show that the proposed method achieves an absolute advantage, particularly on VQA datasets, even when retaining only 1\% of the visual tokens. In contrast, token pruning methods experience a significant performance decline under these conditions. The prefill stage facilitates sufficient interaction between visual and textual information, as the attention mechanism aggregates visual information to key text tokens. The proposed method effectively preserves these key text tokens during KV cache token pruning. Consequently, even when only 1\% of visual tokens are retained, the model maintains good inference performance. In contrast, most token pruning methods do not involve cross-modal interaction, resulting in direct visual information loss that cannot be transferred to the subsequent decoding stage in any form. This leads to insufficient and inaccurate visual guidance for generating, significantly degrading performance.
\begin{table}[]
\centering
\caption{Ablation results on the components of the AirCache.}
\resizebox{1.0\linewidth}{!}{
\begin{tabular}{ccccc}
\hline
Setting & ChatQA \cite{masry2022chartqa} & InfoVQA \cite{mathew2022infographicvqa}& DocVQA \cite{mathew2021docvqa} & TextVQA \cite{singh2019towards} \\ \hline
Continuous window (16) & 70.4 & 56.6 & 61.3 & 55.9\\
Continuous window (32) & 72.9 & 57.8 & 64.1 & 58.2\\
All text tokens & 72.2 & 58.4 & 65.7 & 57.0\\
Visual window (32) & 68.8 & 55.1 & 59.2 & 53.7\\
\midrule
\midrule
Average allocation & 72.2 & 57.5 & 69.9 & 62.4 \\
Pyramid allocation & 69.6 & 54.9 & 55.8 & 52.6 \\
Only w/ $s_t$ & 74.2 & 59.8 & 71.1 & 64.9 \\
Only w/ $s_k$ & 74.7 & 61.4 & 71.9 & 63.6 \\
\midrule
\midrule
Ours & 76.4 & 62.5 & 73.2 & 67.1 \\ \hline
\end{tabular}
}
\vspace{-10pt}
\label{tab: ablation_results}
\end{table}

\subsection{Ablation Studies}
In this section, we evaluate our method under various settings to verify the proposed modules from several perspectives. The base model is LLaVA-OV-7B \cite{li2024llava}.

\noindent \textbf{Elite Observation Window.} We introduce three common types of observation windows to assess the effectiveness of the proposed elite observation window. The first type employs a sequence of continuous text tokens as the observation window for evaluating the significance of visual tokens. The second type utilizes all text tokens, whereas the third employs the continuous visual token from the last visual segment as a reference. As shown in the first column of Table \ref{tab: ablation_results}, the proposed elite observation window yields the best performance compared to the other three aforementioned approaches. The method using only visual tokens performs significantly worse due to the absence of textual instruction guidance, highlighting the importance of activating cross-modal associations in KV cache compression. Additionally, compared to methods that directly use all or portions of continuous text token sequences without considering variations in visual information perception among text tokens, the proposed method selects key text tokens to enhance the consistency of the observation window. This selection ensures greater stability and accuracy in evaluating the significance of visual tokens.

\noindent \textbf{Layer-wise Token Budget Allocation Module.} The second column of Table \ref{tab: ablation_results} presents the comparison using various budget allocation strategies across layers. Aside from the basic equal allocation, we also incorporate the strategy from PyramidKV \cite{cai2024pyramidkv}, which allocates budgets in a descending ratio from shallow to deep layers. The results demonstrate that the proposed budget allocation, based on the strength and skewness of importance scores, delivers superior outcomes. Although PyramidKV \cite{cai2024pyramidkv} perform well in LLMs, its suboptimal results in LVLMs highlight the distinct characteristics of multimodal models and their heightened sensitivity to hierarchical budget allocation. An inappropriate allocation strategy is more likely to adversely affect the final results. We also investigate the separate use of strength and skewness for layer-wise budget quantification, revealing that their combination produces the best results.

\begin{table}[t]
\centering
\caption{Comparison results with different compression audiences, where * indicates uniform compression without distinguishing between visual tokens and text tokens.}
\resizebox{1.0\linewidth}{!}{
\begin{tabular}{ccccc}
\hline
Models & {50\%} & {10\%} & {5\%} & {1\%} \\ \hline
H2O* \cite{zhang2024h2o} & 74.6/407s & 46.5/559s & 32.0/733s & 27.8/1048s\\
H2O \cite{zhang2024h2o} & 79.8/323s & 77.4/251s & 74.0/239s & 71.0/215s \\ \hline
Ours* & 75.8/374s & 53.4/530s & 39.7/688s & 30.1/963s \\
Ours & 80.4/306s & 79.9/235s & 79.4/221s & 76.4/207s \\ \hline
\end{tabular}
}
\label{tab: uniform_compression}
\end{table}

\begin{figure}[t!]
    \centering
    \setlength{\abovecaptionskip}{0cm}
    \setlength{\belowcaptionskip}{-0.5cm}
    \includegraphics[width=0.48\textwidth]{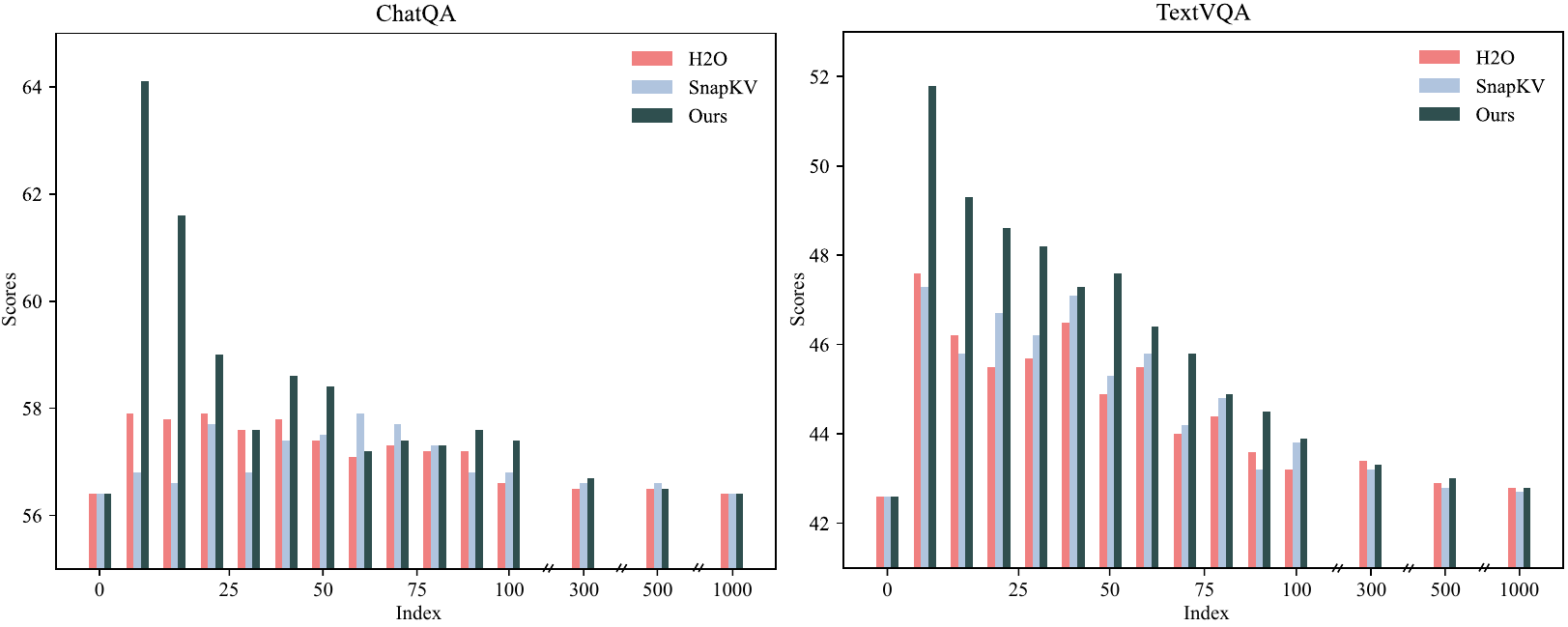}
    \caption{Comparison on ChatQA \cite{masry2022chartqa} and TextVQA \cite{singh2019towards} by retaining only one visual token, which is selected based on the sorting of visual token importance scores using different methods.}
    \vspace{-10pt}
    \label{fig4_one_visual_token}
\end{figure}

\noindent \textbf{Visual Token Importance Evaluation.} Figure \ref{fig4_one_visual_token} visually demonstrates the comparative effectiveness of various methods in evaluating the importance of visual tokens. To directly assess the efficacy of these methods in ranking visual token importance, we removed all non-essential modules and selected only one visual token at a time from the ranked sorting results, ordered from high to low importance. Theoretically, as the importance ranking of tokens declines, the model's performance is also expected to decrease. The graph clearly shows that the proposed method aligns more closely with the theoretical trend, where the visual token ranked as TOP1 achieves the highest metric, followed by a continuous and smooth decline. Notably, after around 10\%, the performance of the visual tokens approximates that of the case with all visual tokens removed. This indicates significant redundancy among visual tokens in the KV cache of LVLMs. The results in Table \ref{tab: main}, which demonstrate that retaining over 10\% of the visual tokens limits performance decline to no more than 1\%, further corroborate this observation. In contrast, the performance derived from other methods exhibit significant noise, indicating instability and deficiencies in their importance assessment.

\noindent \textbf{Unified Compression or Vision-only Compression.} Given that the KV cache in LVLMs encompasses both visual and textual modalities, in contrast to the single textual modality of LLMs, we compared Unified Compression, which does not differentiate between modal differences during KV cache compression, with Vision-only Compression, which compresses only visual tokens. As shown in Table \ref{tab: uniform_compression}, the results indicate that unified compression performs significantly worse than vision-only compression in terms of both inference performance and speed. Moreover, as the compression ratio increases, the inference time for unified compression actually continues to rise. The primary reason is that unified compression disrupts the effective representation of textual instructions, leading the model to produce incorrect outputs and exhibit a tendency for repetition, which in turn increases inference time.

\section{Conclusion}
In this work, we propose Activating Inter-modal Relevancy (AirCache), a novel KV cache compression for efficient LVLMs inference. By leveraging inter-modal relevancy between textual and visual tokens to guide KV cache eviction, we identified significant differences in the attention distribution of visual tokens corresponding to different textual tokens. Based on this finding, we propose an elite observation window with refined key text tokens, which provides a more stable and effective assessment of visual token importance. Furthermore, by exploring the diversity of importance score distribution across different layers, we propose to achieve comprehensive quantification in terms of distribution strength and skewness for adaptive layer-wise compression budget allocation. Extensive experiments across multiple benchmarks and LVLMs demonstrate the effectiveness of AirCache in KV cache compression.

{\small
\bibliographystyle{ieee_fullname}
\bibliography{main}
}

\newpage
\appendix

\section*{Roadmap of Appendix}
In this supplementary material, we first state the limitations of the proposed method and potential future work in Section \ref{sec: A}. Next, we provide more details on the method's application in Section \ref{sec: B}. After that, additional main comparative experiments with more models are discussed in Section \ref{sec: C}. Furthermore, we present additional ablation experiments in Section \ref{sec: D}. Finally, the visualization of chat generation is shown in Section \ref{sec: E}.

\section{Limitations and Future Works}
\label{sec: A}
In performing dynamic allocation of the layer-wise compression budget, the proposed method requires obtaining the strength and skewness of the importance distribution of visual tokens for all layers before determining the allocable budget for each layer. This necessitates storing the complete KV cache after the prefill stage and executing the reduction only once the final compression budget is determined. Consequently, the proposed method is at a disadvantage in terms of peak memory consumption, a challenge also faced by most hierarchical budget allocation methods. Addressing how to maintain a peak memory advantage while supporting dynamic allocation of budgets across layers will be a key focus of our future work. In parallel, we will continue to explore the information flow mechanisms of different modalities in the inference process of LVLMs to further optimize the proposed method.

\section{Implementation Details}
\label{sec: B}
For most methods, we adhere to their initial setup and perform the reduction of the visual KV cache based on the obtained importance ranking of visual tokens and the specified compression ratio. Our findings indicate that merging the dropped KV cache into the KV cache that needs to be retained works effectively on the LLaVA-v1.5 \cite{liu2024improved}. However, this approach tends to cause repetition issues in the LLaVA-OV series \cite{li2024llava}, InternVL2 series \cite{chen2024internvl}, and Qwen2-VL series \cite{wang2024qwen2}, which results in a decline in model performance. Consequently, for Elastic Cache \cite{liu2024efficient}, we omitted the merge operation in the main experiments to achieve optimal performance results. In practical applications, and as observed in most existing multimodal evaluation datasets, visual tokens constitute the majority, while text tokens remain concise and short. The redundancy in the KV cache primarily resides in the visual part. As demonstrated in Table \ref{tab: 8}, a comparison of the actual number of visual tokens and text tokens in these multimodal VQA datasets shows that the visual component accounts for more than 97\%. Thus, compressing only the visual KV cache eliminates redundant cache without affecting the complete expression of text instructions. Unless otherwise specified, all methods and experiments perform KV cache compression solely on the visual part.

\begin{table}[]
\caption{The number of visual tokens and text tokens across different models and evaluation sets.}
\resizebox{1.0\linewidth}{!}{
\begin{tabular}{@{}ccccccccc@{}}
\toprule
\multirow{2}{*}{Models} & \multicolumn{2}{c}{ChatQA \cite{masry2022chartqa}} & \multicolumn{2}{c}{InfoVQA \cite{mathew2022infographicvqa}} & \multicolumn{2}{c}{DocVQA \cite{mathew2021docvqa}} & \multicolumn{2}{c}{TextVQA \cite{singh2019towards}} \\ \cmidrule(l){2-3} \cmidrule(l){4-5} \cmidrule(l){6-7} \cmidrule(l){8-9}
& $N_v$ & $N_t$ & $N_v$ & $N_t$ & $N_v$ & $N_t$ & $N_v$ & $N_t$ \\ \midrule
LLaVA-OV-7B \cite{li2024llava}& 4763 & 47 & 6382 & 45 & 7224 & 42 & 5183 & 39 \\
InternVL2-8B \cite{chen2024internvl}& 1828& 32& 3740& 31& 3230& 28& 1668& 25\\
Qwen2-VL-7B \cite{wang2024qwen2}& 1302& 36& 4450& 34& 4669& 31& 1325& 28\\ \bottomrule
\end{tabular}
}
\label{tab: 8}
\end{table}

\begin{table*}[th]
\caption{The comparison of the KV cache compression methods on multimodal VQA benchmarks. The best result is highlighted in bold.}
\resizebox{1.0\linewidth}{!}{
\begin{tabular}{c|c|cccc|cccc|cccc|cccc}
\hline
    & & \multicolumn{4}{c|}{ChatQA \cite{masry2022chartqa}} & \multicolumn{4}{c|}{InfoVQA \cite{mathew2022infographicvqa}} & \multicolumn{4}{c|}{DocVQA \cite{mathew2021docvqa}} & \multicolumn{4}{c}{TextVQA \cite{singh2019towards}} \\ \cline{3-18} 
    \multirow{-2}{*}{Models} & \multirow{-2}{*}{Methods} & 50\% & 10\% & 5\% & 1\% & 50\% & 10\% & 5\% & 1\% & 50\% & 10\% & 5\% & 1\% & 50\% & 10\% & 5\% & 1\% \\ \hline
    & \cellcolor[HTML]{EFEFEF}Full & \cellcolor[HTML]{EFEFEF}67.9 & \cellcolor[HTML]{EFEFEF}67.9 & \cellcolor[HTML]{EFEFEF}67.9 & \cellcolor[HTML]{EFEFEF}67.9 & \cellcolor[HTML]{EFEFEF}50.1 & \cellcolor[HTML]{EFEFEF}50.1 & \cellcolor[HTML]{EFEFEF}50.1 & \cellcolor[HTML]{EFEFEF}50.1 & \cellcolor[HTML]{EFEFEF}80.0 & \cellcolor[HTML]{EFEFEF}80.0 & \cellcolor[HTML]{EFEFEF}80.0 & \cellcolor[HTML]{EFEFEF}80.0 & \cellcolor[HTML]{EFEFEF}70.8 & \cellcolor[HTML]{EFEFEF}70.8 & \cellcolor[HTML]{EFEFEF}70.8 & \cellcolor[HTML]{EFEFEF}70.8 \\
     & H2O \cite{zhang2024h2o} & 
     67.7&62.0&57.7&53.5&50.0&42.7&38.8&33.7&79.8&73.8&68.2&57.0&70.1&58.4&52.3&47.6\\
     & Elastic \cite{liu2024efficient} & 
     67.5&61.8&57.6&54.1&50.1&42.6&39.7&32.2&79.8&74.0&68.8&57.5&70.3&59.8&54.7&47.2\\
     & PrefixKV \cite{wang2024prefixkv} & 
     \bf 67.9&62.1&58.0&53.3&49.9&43.8&40.6&34.2&79.7&74.3&70.4&59.6&70.1&60.7&55.3&48.6\\
     & SnapKV \cite{li2024snapkv} &  
     67.8&63.7&59.8&56.7&50.1&47.3&44.4&39.7&79.8&76.5&72.1&61.2&70.4&64.5&60.7&52.2\\
    \multirow{-6}{*}{InternVL2-1B \cite{chen2024internvl}}  & Ours &
    67.8&\bf65.7&\bf63.5&\bf60.8&\bf50.1&\bf49.6&\bf47.5&\bf45.8&\bf80.0&\bf77.7&\bf74.3&\bf68.5&\bf70.6&\bf68.9&\bf66.5&\bf59.3\\ \hline \hline 
    & \cellcolor[HTML]{EFEFEF}Full & \cellcolor[HTML]{EFEFEF}81.1 & \cellcolor[HTML]{EFEFEF}81.1 & \cellcolor[HTML]{EFEFEF}81.1 & \cellcolor[HTML]{EFEFEF}81.1 & \cellcolor[HTML]{EFEFEF}65.9 & \cellcolor[HTML]{EFEFEF}65.9 & \cellcolor[HTML]{EFEFEF}65.9 & \cellcolor[HTML]{EFEFEF}65.9 & \cellcolor[HTML]{EFEFEF}88.1 & \cellcolor[HTML]{EFEFEF}88.1 & \cellcolor[HTML]{EFEFEF}88.1 & \cellcolor[HTML]{EFEFEF}88.1 & \cellcolor[HTML]{EFEFEF}74.7 & \cellcolor[HTML]{EFEFEF}74.7 & \cellcolor[HTML]{EFEFEF}74.7 & \cellcolor[HTML]{EFEFEF}74.7 \\
     & H2O \cite{zhang2024h2o} &
     81.1&79.2&77.6&72.1&65.9&61.1&57.4&51.8&79.9&80.1&75.4&69.2&74.2&66.4&58.8&51.3\\
     & Elastic \cite{liu2024efficient} &
     81.1&79.4&77.9&73.6&65.8&61.8&59.2&53.3&79.7&80.7&75.9&69.6&74.0&67.0&60.4&52.6\\
     & PrefixKV \cite{wang2024prefixkv} &
     81.0&79.5&77.8&73.2&65.9&62.6&59.1&53.4&88.0&81.4&76.4&70.3&74.4&67.5&61.4&53.7\\
     & SnapKV \cite{li2024snapkv} &  
     81.1&79.3&78.5&74.6&65.9&64.3&61.8&56.7&88.0&84.3&79.7&73.2&74.4&70.3&65.8&60.3\\
    \multirow{-6}{*}{InternVL2-4B \cite{chen2024internvl}} & Ours & 
    \bf81.1&\bf80.4&\bf79.6&\bf77.7&\bf66.0&\bf65.5&\bf64.2&\bf62.1&\bf88.1&\bf86.8&\bf84.3&\bf81.5&\bf74.5&\bf73.6&\bf70.7&\bf67.4\\ \hline \hline
     & \cellcolor[HTML]{EFEFEF}Full & \cellcolor[HTML]{EFEFEF}85.4 & \cellcolor[HTML]{EFEFEF}85.4 & \cellcolor[HTML]{EFEFEF}85.4 & \cellcolor[HTML]{EFEFEF}85.4 & \cellcolor[HTML]{EFEFEF}75.4 & \cellcolor[HTML]{EFEFEF}75.4 & \cellcolor[HTML]{EFEFEF}75.4 & \cellcolor[HTML]{EFEFEF}75.4 & \cellcolor[HTML]{EFEFEF}92.1 & \cellcolor[HTML]{EFEFEF}92.1 & \cellcolor[HTML]{EFEFEF}92.1 & \multicolumn{1}{l|}{\cellcolor[HTML]{EFEFEF}92.1} & \multicolumn{1}{l}{\cellcolor[HTML]{EFEFEF}82.5} & \multicolumn{1}{l}{\cellcolor[HTML]{EFEFEF}82.5} & \multicolumn{1}{l}{\cellcolor[HTML]{EFEFEF}82.5} & \multicolumn{1}{l}{\cellcolor[HTML]{EFEFEF}82.5} \\
     & H2O \cite{zhang2024h2o} & 
     84.9&82.4&80.4&78.6&75.0&71.8&65.1&62.5&91.9&84.4&81.6&75.1&82.3&75.2&70.3&65.2\\
     & Elastic \cite{liu2024efficient} & 
     84.6&82.8&81.6&78.3&74.8&73.6&65.5&62.7&91.8&83.8&81.2&74.3&82.4&75.6&70.7&65.7\\
     & PrefixKV \cite{wang2024prefixkv} &
     84.8&82.2&81.5&78.8&75.2&73.8&66.4&63.2&92.0&84.2&81.5&74.7&82.4&75.5&70.6&65.4\\
     & SnapKV \cite{li2024snapkv} &85.3&83.5&83.0&80.1&75.4&74.1&69.5&65.6&91.9&86.6&86.3&82.5&\bf82.5&78.6&74.3&71.1 \\
    \multirow{-6}{*}{InternVL2-26B \cite{chen2024internvl}} & Ours&\bf85.5&\bf84.7&\bf84.2&\bf82.3&\bf75.4&\bf74.8&\bf72.8&\bf70.7&\bf92.1&\bf91.4&\bf89.0&\bf86.8&82.4&\bf81.7&\bf78.2&\bf76.6 \\ \hline
\end{tabular}
}
\label{tab: 9}
\vspace{-10pt}
\end{table*}

\begin{table*}[]
\caption{Quantitative results on inference latency and throughput. The number of tokens output is consistently set to 512.}
\resizebox{1.0\linewidth}{!}{
\begin{tabular}{@{}c|cccccccccc@{}}
\toprule
 & & & \multicolumn{2}{c}{Prefill Latency (s)} & \multicolumn{3}{c}{Decoding Latency (s)}                 & \multicolumn{3}{c}{Throughput (token/s)} \\ \cmidrule(l){4-11} 
\multirow{-2}{*}{Models} & \multirow{-2}{*}{Batch Size} & \multirow{-2}{*}{Prompt Length} & Full        & Ours & Full & 50\% & 10\% & Full & 50\% & 10\% \\ \midrule
\multicolumn{1}{c|}{} & & 2k & \cellcolor[HTML]{EFEFEF}1.2 & 1.4\(_{\color{blue}{\tiny+16.7\%}}\) & \cellcolor[HTML]{EFEFEF}9.3  & 7.0\(_{\color{blue}{\tiny+24.7\%}}\)  & 5.9\(_{\color{blue}{\tiny+36.6\%}}\)  & \cellcolor[HTML]{EFEFEF}440 & 585\(_{\color{blue}{\tiny+33.0\%}}\)  & 694\(_{\color{blue}{\tiny+36.6\%}}\)  \\
\multicolumn{1}{c|}{} & & 8k & \cellcolor[HTML]{EFEFEF}4.6  & 5.0\(_{\color{blue}{\tiny+8.7\%}}\)  & \cellcolor[HTML]{EFEFEF}13.3 & 10.5\(_{\color{blue}{\tiny+21.1\%}}\) & 8.8\(_{\color{blue}{\tiny+33.8\%}}\)  & \cellcolor[HTML]{EFEFEF}308 & 390\(_{\color{blue}{\tiny+26.6\%}}\)  & 465\(_{\color{blue}{\tiny+51.0\%}}\)  \\
\multicolumn{1}{c|}{} & & 16k & \cellcolor[HTML]{EFEFEF}9.8  & 10.5\(_{\color{blue}{\tiny+7.1\%}}\) & \cellcolor[HTML]{EFEFEF}24.8 & 15.7\(_{\color{blue}{\tiny+36.7\%}}\) & 11.9\(_{\color{blue}{\tiny+52.0\%}}\) & \cellcolor[HTML]{EFEFEF}165 & 261\(_{\color{blue}{\tiny+58.2\%}}\)  & 344\(_{\color{blue}{\tiny+52.0\%}}\)  \\
\multicolumn{1}{c|}{} & \multirow{-4}{*}{8} & 32k & \cellcolor[HTML]{EFEFEF}23.8 & 24.8\(_{\color{blue}{\tiny+4.2\%}}\) & \cellcolor[HTML]{EFEFEF}46.2 & 28.1\(_{\color{blue}{\tiny+39.2\%}}\) & 18.2\(_{\color{blue}{\tiny+60.6\%}}\) & \cellcolor[HTML]{EFEFEF}89  & 146\(_{\color{blue}{\tiny+64.0\%}}\)  & 225\(_{\color{blue}{\tiny+60.4\%}}\)  \\ \cmidrule(l){2-11} 
\multicolumn{1}{c|}{} & & 2k & \cellcolor[HTML]{EFEFEF}2.5  & 2.7\(_{\color{blue}{\tiny+8.0\%}}\)  & \cellcolor[HTML]{EFEFEF}12.2 & 7.4\(_{\color{blue}{\tiny+39.3\%}}\)  & 6.0\(_{\color{blue}{\tiny+50.8\%}}\)  & \cellcolor[HTML]{EFEFEF}671 & 1107\(_{\color{blue}{\tiny+65.0\%}}\) & 1365\(_{\color{blue}{\tiny+50.8\%}}\) \\
\multicolumn{1}{c|}{} & & 8k & \cellcolor[HTML]{EFEFEF}9.9  & 10.6\(_{\color{blue}{\tiny+7.1\%}}\) & \cellcolor[HTML]{EFEFEF}21.6 & 12.9\(_{\color{blue}{\tiny+40.3\%}}\) & 8.9\(_{\color{blue}{\tiny+58.8\%}}\)  & \cellcolor[HTML]{EFEFEF}379 & 635\(_{\color{blue}{\tiny+67.5\%}}\)  & 920\(_{\color{blue}{\tiny+58.8\%}}\)  \\
\multicolumn{1}{c|}{\multirow{-7}{*}{InternVL2-8B \cite{chen2024internvl}}} & \multirow{-3}{*}{16} & 16k & \cellcolor[HTML]{EFEFEF}21.6 & 22.4\(_{\color{blue}{\tiny+3.7\%}}\) & \cellcolor[HTML]{EFEFEF}28.5 & 16.2\(_{\color{blue}{\tiny+43.2\%}}\) & 10.0\(_{\color{blue}{\tiny+64.9\%}}\) & \cellcolor[HTML]{EFEFEF}287 & 506\(_{\color{blue}{\tiny+76.3\%}}\)  & 819\(_{\color{blue}{\tiny+65.0\%}}\)  \\ \midrule
\multicolumn{1}{c|}{} & & 2k & \cellcolor[HTML]{EFEFEF}1.1  & 1.3\(_{\color{blue}{\tiny+18.2\%}}\) & \cellcolor[HTML]{EFEFEF}8.4  & 6.6\(_{\color{blue}{\tiny+21.4\%}}\)  & 5.2\(_{\color{blue}{\tiny+38.1\%}}\)  & \cellcolor[HTML]{EFEFEF}488 & 621\(_{\color{blue}{\tiny+27.3\%}}\)  & 788\(_{\color{blue}{\tiny+38.1\%}}\)  \\
\multicolumn{1}{c|}{} & & 8k & \cellcolor[HTML]{EFEFEF}4.3  & 4.7\(_{\color{blue}{\tiny+27.3\%}}\)  & \cellcolor[HTML]{EFEFEF}12.6 & 9.7\(_{\color{blue}{\tiny+23.0\%}}\)  & 8.1\(_{\color{blue}{\tiny+35.7\%}}\)  & \cellcolor[HTML]{EFEFEF}325 & 422\(_{\color{blue}{\tiny+29.8\%}}\)  & 506\(_{\color{blue}{\tiny+55.7\%}}\)  \\
\multicolumn{1}{c|}{} & & 16k & \cellcolor[HTML]{EFEFEF}9.4  & 10.2\(_{\color{blue}{\tiny+8.5\%}}\) & \cellcolor[HTML]{EFEFEF}23.7 & 14.9\(_{\color{blue}{\tiny+37.1\%}}\) & 11.2\(_{\color{blue}{\tiny+52.7\%}}\) & \cellcolor[HTML]{EFEFEF}173 & 275\(_{\color{blue}{\tiny+59.0\%}}\)  & 366\(_{\color{blue}{\tiny+112.9\%}}\) \\
\multicolumn{1}{c|}{} & \multirow{-4}{*}{8} & 32k & \cellcolor[HTML]{EFEFEF}22.7 & 24.0\(_{\color{blue}{\tiny+5.7\%}}\) & \cellcolor[HTML]{EFEFEF}45.0 & 26.2\(_{\color{blue}{\tiny+41.8\%}}\) & 14.6\(_{\color{blue}{\tiny+67.6\%}}\) & \cellcolor[HTML]{EFEFEF}91  & 156\(_{\color{blue}{\tiny+71.4\%}}\)  & 214\(_{\color{blue}{\tiny+135.2\%}}\) \\ \cmidrule(l){2-11} 
\multicolumn{1}{c|}{} & & 2k & \cellcolor[HTML]{EFEFEF}2.1  & 2.3\(_{\color{blue}{\tiny+9.5\%}}\)  & \cellcolor[HTML]{EFEFEF}10.1 & 6.8\(_{\color{blue}{\tiny+32.7\%}}\)  & 5.2\(_{\color{blue}{\tiny+48.5\%}}\)  & \cellcolor[HTML]{EFEFEF}811 & 1205\(_{\color{blue}{\tiny+48.6\%}}\) & 1575\(_{\color{blue}{\tiny+94.2\%}}\) \\
\multicolumn{1}{c|}{} & & 8k & \cellcolor[HTML]{EFEFEF}8.6  & 9.2\(_{\color{blue}{\tiny+7.0\%}}\)  & \cellcolor[HTML]{EFEFEF}19.7 & 12.1\(_{\color{blue}{\tiny+38.6\%}}\) & 8.4\(_{\color{blue}{\tiny+57.4\%}}\)  & \cellcolor[HTML]{EFEFEF}426 & 677\(_{\color{blue}{\tiny+37.1\%}}\)  & 975\(_{\color{blue}{\tiny+128.9\%}}\) \\
\multicolumn{1}{c|}{\multirow{-7}{*}{Qwen2-VL-7B \cite{wang2024qwen2}}}  & \multirow{-3}{*}{16} & 16k & \cellcolor[HTML]{EFEFEF}19.2 & 20.1\(_{\color{blue}{\tiny+4.7\%}}\) & \cellcolor[HTML]{EFEFEF}27.0 & 15.6\(_{\color{blue}{\tiny+42.2\%}}\) & 9.7\(_{\color{blue}{\tiny+64.1\%}}\)  & \cellcolor[HTML]{EFEFEF}303 & 525\(_{\color{blue}{\tiny+73.3\%}}\)  & 845\(_{\color{blue}{\tiny+179.0\%}}\)   \\ \bottomrule
\end{tabular}
}
\label{tab: 10}
\end{table*}

\section{Additional Main Results}
\label{sec: C}
\noindent \textbf{Comparison with various model parameter sizes.} Table \ref{tab: 9} displays the comparison results of different parameter-sized InternVL2 \cite{chen2024internvl} series models across various VQA datasets as the compression ratio varies. Similar to the conclusions drawn from experiments with different model architectures, the proposed method achieves superior results on models with different parameter sizes compared to existing methods. For instance, when retaining only 1\% of the visual KV cache, the proposed method outperforms the SnapKV \cite{li2024snapkv} method by an average of approximately 4.3\% to 6.0\% across four VQA evaluation datasets as the model parameter size varies. By synthesizing experiments on different architectures and base models with varying parameter sizes, we observe that the proposed method not only achieves better compression results but also demonstrates good general applicability. Furthermore, a comparison of models with different parameter sizes reveals that as the model parameter size decreases, the impact of KV cache compression on model performance becomes more significant. This trend indicates that smaller parameter-sized models are less effective at integrating information within tokens, thereby placing a greater emphasis on the KV cache compression method's ability to select important visual tokens. The proposed method demonstrates a superior capability in assessing the importance of visual tokens, thereby more effectively reducing model performance loss.

\noindent \textbf{Inference efficiency on more LVLMs.} Inference efficiency on more LVLMs. Table \ref{tab: 10} further presents the comparison of inference latency between the proposed method and the full cache on InternVL2-8B \cite{chen2024internvl} and Qwen2-VL-7B \cite{wang2024qwen2}. From the table, it can be observed that when the input demand is relatively low, the model's need for the KV cache is reduced, thus limiting the gains from the KV cache compression method. Nevertheless, there is at least a 21\% reduction in decoding latency and a 27\% increase in throughput. As the input demand continues to rise, the benefits from KV cache compression become more significant. For example, in the case of the Qwen2-VL-7B \cite{wang2024qwen2} with a batch size of 16 and a prompt length of 16k, the proposed method can reduce decoding latency by 42\% and increase throughput by 73\% with almost no impact on model performance, while only adding 5\% to prefill latency.

\begin{table*}[]
\caption{The detailed comparison of the KV cache compression methods on MMBench-Video \cite{fang2024mmbench}. CP (coarse perception), FP-S (single-instance fine-grained perception), FP-C (cross-instance fine-grained perception), HL (Hallucination), LR (logic reasoning), AR (attribute reasoning), RR (relation reasoning), CSR (commonsense reasoning), TR (temporal reasoning).}
\resizebox{1.02\linewidth}{!}{
\begin{tabular}{|c|ccccccccccccc|}
\hline
& Methods & Overall & Perception & Reasoning & CP & FP-S & FP-C & HL & LR & AR & RR & CSR & TR \\ \cline{2-14} 
\multirow{-2}{*}{Ratio} & \cellcolor[HTML]{EFEFEF}Full & \cellcolor[HTML]{EFEFEF}1.81 & \cellcolor[HTML]{EFEFEF}1.86 & \cellcolor[HTML]{EFEFEF}1.70 & \cellcolor[HTML]{EFEFEF}1.90 & \cellcolor[HTML]{EFEFEF}1.94 & \cellcolor[HTML]{EFEFEF}1.70 & \cellcolor[HTML]{EFEFEF}0.81 & \cellcolor[HTML]{EFEFEF}1.63 & \cellcolor[HTML]{EFEFEF}1.84 & \cellcolor[HTML]{EFEFEF}1.64 & \cellcolor[HTML]{EFEFEF}1.85 & \cellcolor[HTML]{EFEFEF}1.57 \\ \hline
& H2O \cite{zhang2024h2o} & 1.66 & 1.68 & 1.56 & 1.76 & 1.74 & 1.66 & 0.71 & 1.56 & 1.76 & 1.65 & 1.63 & 1.42 \\
& Elastic \cite{liu2024efficient} & 1.68 & 1.71 & 1.60 & 1.79 & 1.75 & 1.66 & 0.73 & 1.58 & 1.77 & 1.67 & 1.66 & 1.44 \\
& PrefixKV \cite{wang2024prefixkv} & 1.72 & 1.75 & 1.63 & 1.84 & 1.79 & 1.68 & 0.82 & 1.58 & 1.79 & 1.68 & 1.68 & 1.48 \\
& SnapKV \cite{li2024snapkv} & 1.75 & 1.80 & 1.67 & 1.88 & 1.88 & 1.59 & 0.77 & 1.58 & 1.83 & 1.73 & 1.85 & 1.47 \\
\multirow{-5}{*}{50\%} & Ours & 1.80 & 1.84 & 1.69 & 1.89 & 1.94 & 1.66 & 0.81 & 1.61 & 1.84 & 1.77 & 1.79 & 1.53 \\ \hline
& H2O \cite{zhang2024h2o} & 1.63 & 1.64 & 1.61 & 1.72 & 1.71 & 1.52 & 0.74 & 1.54 & 1.81 & 1.64 & 1.73 & 1.43 \\
& Elastic \cite{liu2024efficient} & 1.62 & 1.64 & 1.58 & 1.77 & 1.67 & 1.51 & 0.79 & 1.46 & 1.71 & 1.61 & 1.68 & 1.48 \\
& PrefixKV \cite{wang2024prefixkv} & 1.68 & 1.74 & 1.58 & 1.81 & 1.73 & 1.63 & 0.80 & 1.56 & 1.79 & 1.68 & 1.69 & 1.47 \\
& SnapKV \cite{li2024snapkv} & 1.70 & 1.76 & 1.60 & 1.83 & 1.83 & 1.62 & 0.82 & 1.55 & 1.70 & 1.67 & 1.78 & 1.44 \\
\multirow{-5}{*}{10\%} & Ours & 1.78 & 1.80 & 1.65 & 1.85 & 1.91 & 1.64 & 0.82 & 1.58 & 1.80 & 1.75 & 1.79 & 1.48 \\ \hline
& H2O \cite{zhang2024h2o} & 1.47 & 1.45 & 1.48 & 1.62 & 1.51 & 1.40 & 0.72 & 1.20 & 1.51 & 1.45 & 1.55 & 1.42 \\
& Elastic \cite{liu2024efficient} & 1.51 & 1.52 & 1.51 & 1.65 & 1.54 & 1.40 & 0.75 & 1.22 & 1.54 & 1.50 & 1.60 & 1.45 \\
& PrefixKV \cite{wang2024prefixkv} & 1.50 & 1.48 & 1.50 & 1.63 & 1.52 & 1.39 & 0.74 & 1.20 & 1.51 & 1.48 & 1.59 & 1.44 \\
& SnapKV \cite{li2024snapkv} & 1.55 & 1.56 & 1.54 & 1.70 & 1.60 & 1.45 & 0.82 & 1.30 & 1.63 & 1.57 & 1.69 & 1.50 \\
\multirow{-5}{*}{1\%} & Ours & 1.67 & 1.70 & 1.58 & 1.78 & 1.76 & 1.65 & 0.72 & 1.58 & 1.76 & 1.65 & 1.64 & 1.52 \\ \hline
\end{tabular}
}
\label{tab: 11}
\end{table*}

\begin{table}[]
\caption{The results that the dropped KV cache is merged with the nearest preserved KV cache at different proportions. 100\% means complete merging is used, while 0\% means complete dropping is used.}
\resizebox{1.02\linewidth}{!}{
\begin{tabular}{@{}ccccccccc@{}}
\toprule
Datasets                 & Ratio & 100\% & 80\% & 60\% & 40\% & 20\% & 0\% \\ \midrule
\multirow{2}{*}{ChatQA \cite{masry2022chartqa}}  & 10\% & 76.4 & 77.0 & 77.4 & 78.6 & 79.2 & 79.9 & \\
                         & 1\% & 72.2 & 73.6 & 74.2 & 75.5 & 76.1 & 76.4 & \\ \hline \hline
\multirow{2}{*}{InfoVQA \cite{mathew2022infographicvqa}} & 10\% & 61.1 & 63.7 & 64.5 & 64.9 & 65.5 & 65.7 & \\
                         & 1\% & 56.4 & 57.6 & 58.8 & 60.4 & 61.8 & 62.5 & \\ \hline \hline
\multirow{2}{*}{DocVQA \cite{mathew2021docvqa}}  & 10\% & 80.3 & 81.6 & 82.7 & 83.8 & 84.3 & 85.5 & \\
                         & 1\% & 67.0 & 68.9 & 70.5 & 71.8 & 72.6 & 73.2 & \\ \hline \hline
\multirow{2}{*}{TextVQA \cite{singh2019towards}} & 10\% & 68.8 & 70.6 & 72.2 & 73.7 & 74.6 & 75.3 & \\
                         & 1\% & 59.7 & 61.8 & 63.2 & 65.3 & 66.6 & 67.1 & \\ \bottomrule
\end{tabular}
}
\label{tab: 12}
\end{table}

\begin{table}[]
\caption{Comparison of results across different evaluation sets and compression ratios with varying relevance thresholds.}
\resizebox{1.02\linewidth}{!}{
\begin{tabular}{@{}ccccccccccc@{}}
\toprule
Datasets                 & Ratio & 0.99 & 0.95 & 0.9 & 0.85 & 0.8 & 0.75 & 0.7 & 0.65 & 0.6 \\ \midrule
\multirow{2}{*}{ChatQA \cite{masry2022chartqa}}  & 10\% & 78.4 & 79.8 & 79.9 & 79.6 & 78.7 & 78.6 & 78.1 & 77.5 & 77.7 \\
                         & 1\% & 75.8 & 76.5 & 76.4 & 76.2 & 75.5 & 74.1 & 73.3 & 72.9 & 71.5 \\ \hline \hline
\multirow{2}{*}{InfoVQA \cite{mathew2022infographicvqa}} & 10\% & 65.1 & 65.5 & 65.7 & 65.7 & 65.0 & 63.7 & 62.2 & 61.6 & 59.4 \\
                         & 1\% & 61.9 & 62.3 & 62.5 & 62.4 & 61.2 & 58.6 & 55.4 & 53.7 & 52.2 \\ \hline \hline
\multirow{2}{*}{DocVQA \cite{mathew2021docvqa}}  & 10\% & 82.4 & 85.6 & 85.5 & 84.7 & 81.3 & 78.8 & 76.1 & 75.5 & 74.8 \\
                         & 1\% & 71.2 & 72.8 & 73.2 & 72.9 & 70.0 & 65.5 & 61.8 & 58.9 & 55.6 \\ \hline \hline
\multirow{2}{*}{TextVQA \cite{singh2019towards}} & 10\% & 73.4 & 75.0 & 75.3 & 75.4 & 74.6 & 74.0 & 73.3 & 72.5 & 71.4 \\
                         & 1\% & 65.5 & 67.2 & 67.1 & 66.8 & 65.2 & 64.7 & 63.5 & 62.4 & 61.2 \\ \bottomrule
\end{tabular}
}
\label{tab: 13}
\end{table}

\begin{table}[]
\caption{Layer-wise budget Jensen-Shannon divergence across different datasets and compression ratios, where Avg. corresponds to using the average allocation strategy.}
\resizebox{1.02\linewidth}{!}{
\begin{tabular}{@{}cccccc@{}}
\toprule
Ratio                 & Setting      & ChatQA \cite{masry2022chartqa} & InfoVQA \cite{mathew2022infographicvqa} & DocVQA \cite{mathew2021docvqa} & TextVQA \cite{singh2019towards} \\ \midrule
\multirow{3}{*}{10\%} & Avg. \& $s_t$ & 0.46 & 0.54 & 0.61 & 0.57 \\
                      & Avg. \& $s_k$ & 0.53 & 0.49 & 0.58 & 0.55 \\
                      & $s_t$ \& $s_k$ & 0.62 & 0.58 & 0.63 & 0.52 \\ \hline \hline
\multirow{3}{*}{1\%}  & Avg. \& $s_t$ & 0.51 & 0.58 & 0.63 & 0.59 \\
                      & Avg. \& $s_k$ & 0.57 & 0.52 & 0.60 & 0.58 \\
                      & $s_t$ \& $s_k$ & 0.68 & 0.61 & 0.59 & 0.57 \\ \bottomrule
\end{tabular}
}
\label{tab: 14}
\end{table}

\begin{figure}[t!]
    \centering
    \setlength{\abovecaptionskip}{0cm}
    \setlength{\belowcaptionskip}{-0.5cm}
    \includegraphics[width=0.48\textwidth]{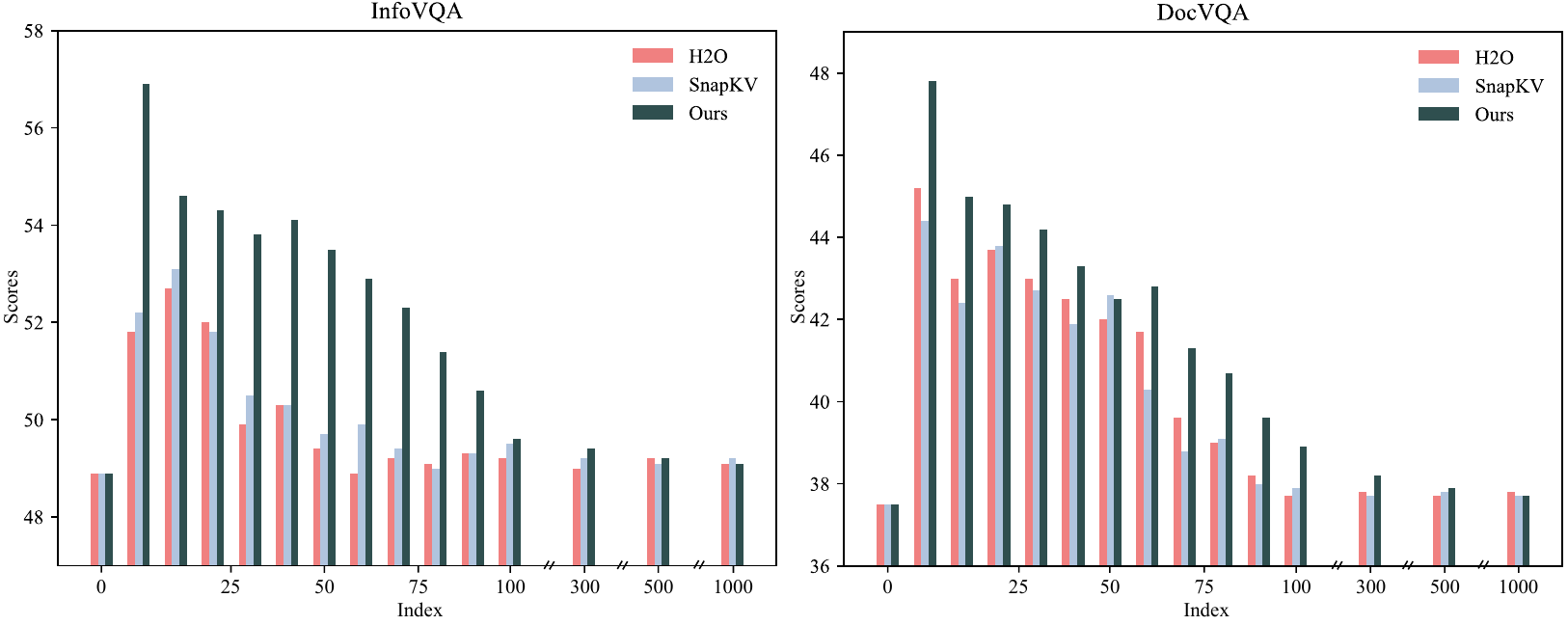}
    \caption{Comparison on InfoVQA \cite{mathew2022infographicvqa} and DocVQA \cite{mathew2021docvqa} by retaining only one visual token, which is selected based on the sorting of visual token importance scores using different methods.}
    \vspace{-10pt}
    \label{fig6_one_visual_token}
\end{figure}

\noindent \textbf{Detailed results of MMBench-Video.} Table \ref{tab: 11} presents the breakdown scores of different methods applied to the LLaVA-OV-7B \cite{li2024llava} on the MMBench-Video \cite{fang2024mmbench} evaluation dataset. The proposed method outperforms existing methods in most subcategories, demonstrating its superior performance and stability. Notably, as the compression ratio increases, the advantages of the proposed method become more pronounced, especially for perceptual items that are more sensitive to visual information. By more accurately assessing the importance of visual tokens, the proposed method retains the critical visual KV cache, thereby minimizing the loss of model performance.

\section{Additional Ablation Studies}
\label{sec: D}
\noindent \textbf{KV Cache Merge \emph{vs.} KV Cache Drop.} Table \ref{tab: 12} compares the model performance using merge and drop strategies for KV cache compression. The drop strategy clearly outperforms the merge strategy, with performance decline becoming more significant as the merge proportion increases. This phenomenon persists across different VQA evaluation sets and various compression ratios, indicating that directly dropping less important visual KV cache in LVLMs is a wiser choice. While the drop operation results in loss of visual information in the corresponding KV cache, the full token attention interaction during the prefill stage allows the remaining important visual tokens and text tokens to potentially absorb this missing information. This absorption helps mitigate the information loss caused by dropping KV cache during subsequent decoding. Conversely, although the merge operation appears to preserve all visual information, the model lacks the ability to decode the original visual information from the merged visual KV cache. This operation may disrupt the representation of important visual information, ultimately leading to a decline in model performance.

\noindent \textbf{Relevance Threshold $\alpha$.} Table \ref{tab: 13} compares model performance under different relevance threshold. A relevance threshold around 0.9 achieves the best overall performance across various evaluation sets and compression ratios. If the relevance threshold is set too high or too low, it can lead to incomplete expression of instruction information or the inclusion of noise, respectively. This degrades the quality of visual KV cache importance assessment, thereby affecting the model's performance after KV cache compression. Comparing a higher relevance threshold with a lower one reveals that introducing more noisy text significantly affects model performance. This emphasizes the importance of filtering out irrelevant text tokens within the observation window when compressing KV cache in LVLMs.

\noindent \textbf{The Consistency of Strength and Skewness.} To illustrate the difference between the dynamic budget and the average budget derived from the distribution strength and skewness used in this method, we recorded the budget distribution differences across various evaluation instances. The differences are quantified using Jensen-Shannon (JS) divergence, which ranges from 0 to 1. A JS divergence closer to 0 indicates smaller differences between the two distributions, while a value closer to 1 indicates larger differences. As shown in Table \ref{tab: 14}, the hierarchical budgets allocated based solely on the strength and skewness of the importance distribution are similar to those with an average allocation, indicating a complementary relationship. The former examines the layer's emphasis on visual information, while the latter focuses on the layer's ability to understand and interpret visual information. Combining both approaches can lead to better model performance.

\noindent \textbf{Ablation Results of the Visual KV Cache Importance Evaluation.} Figure \ref{fig6_one_visual_token} shows a comparison of selecting a single visual token based on the importance ranking of visual KV cache obtained by various methods on InfoVQA \cite{mathew2022infographicvqa} and DocVQA \cite{mathew2021docvqa}. As the importance of the selected visual token decreases, the performance of our proposed method also decreases reasonably. Additionally, for the same importance ranking, the performance of our proposed method is superior to that of existing methods.

\section{Visualization of Chat Generation}
\label{sec: E}
\begin{figure}[t!]
    \centering
    \setlength{\abovecaptionskip}{0cm}
    \setlength{\belowcaptionskip}{-0.5cm}
    \includegraphics[width=0.48\textwidth]{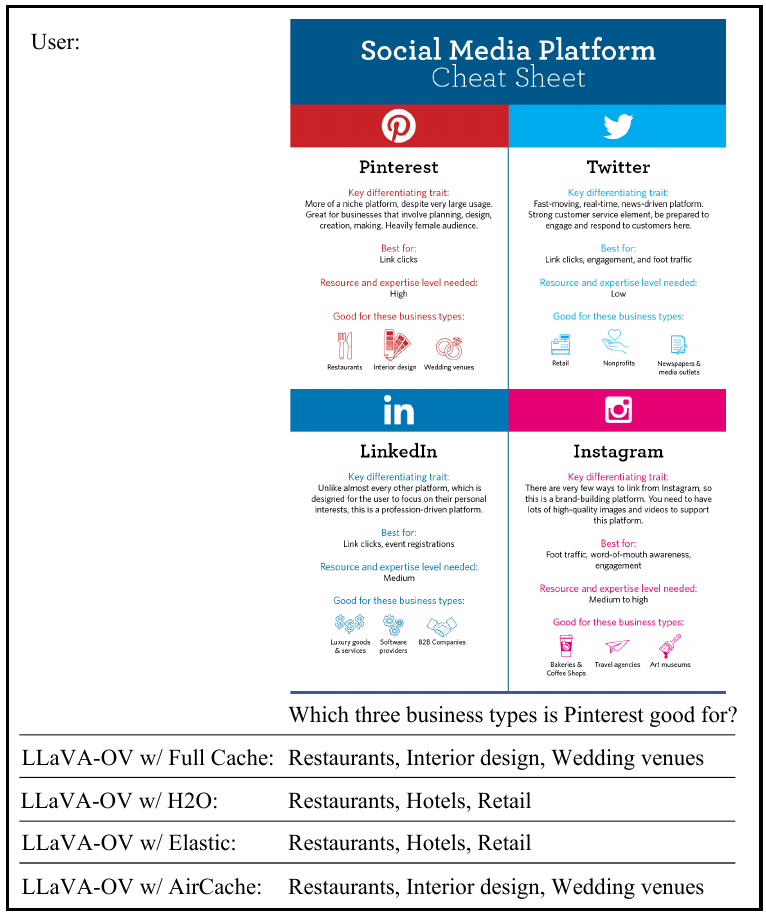}
    \caption{Chat example applying KV cache compression methods on LLAVA-OV-7B \cite{li2024llava}.}
    \vspace{-10pt}
    \label{fig7: chat_generation_e1}
\end{figure}

\begin{figure}[t!]
    \centering
    \setlength{\abovecaptionskip}{0cm}
    \setlength{\belowcaptionskip}{-0.5cm}
    \includegraphics[width=0.48\textwidth]{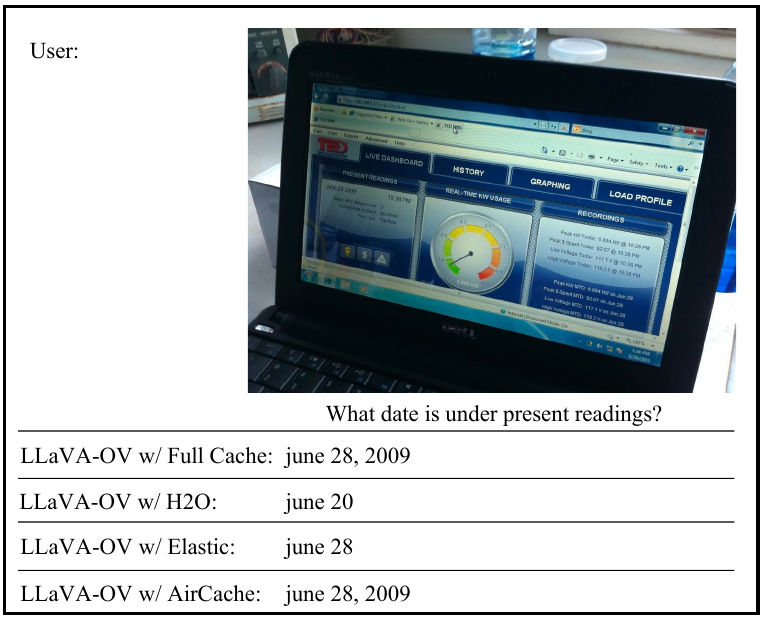}
    \caption{Chat example applying KV cache compression methods on LLAVA-OV-7B \cite{li2024llava}.}
    \vspace{-10pt}
    \label{fig8: chat_generation_e2}
\end{figure}

\begin{figure}[t!]
    \centering
    \setlength{\abovecaptionskip}{0cm}
    \setlength{\belowcaptionskip}{-0.5cm}
    \includegraphics[width=0.48\textwidth]{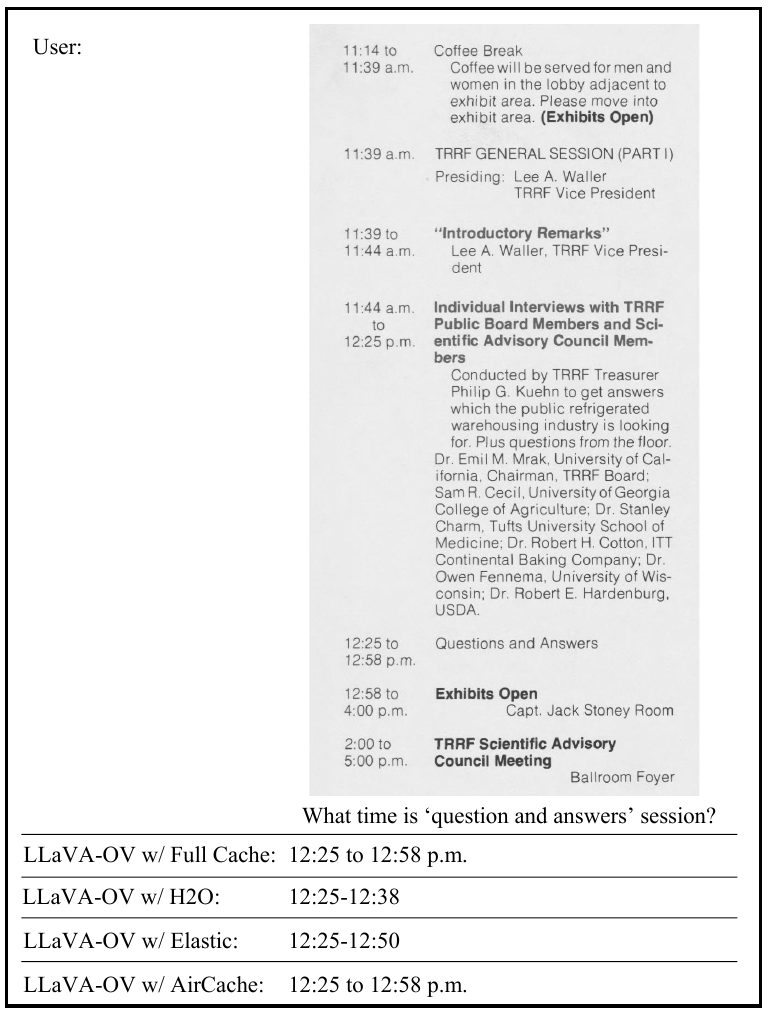}
    \caption{Chat example applying KV cache compression methods on LLAVA-OV-7B \cite{li2024llava}.}
    \vspace{-10pt}
    \label{fig9: chat_generation_e3}
\end{figure}

\begin{figure}[t!]
    \centering
    \setlength{\abovecaptionskip}{0cm}
    \setlength{\belowcaptionskip}{-0.5cm}
    \includegraphics[width=0.48\textwidth]{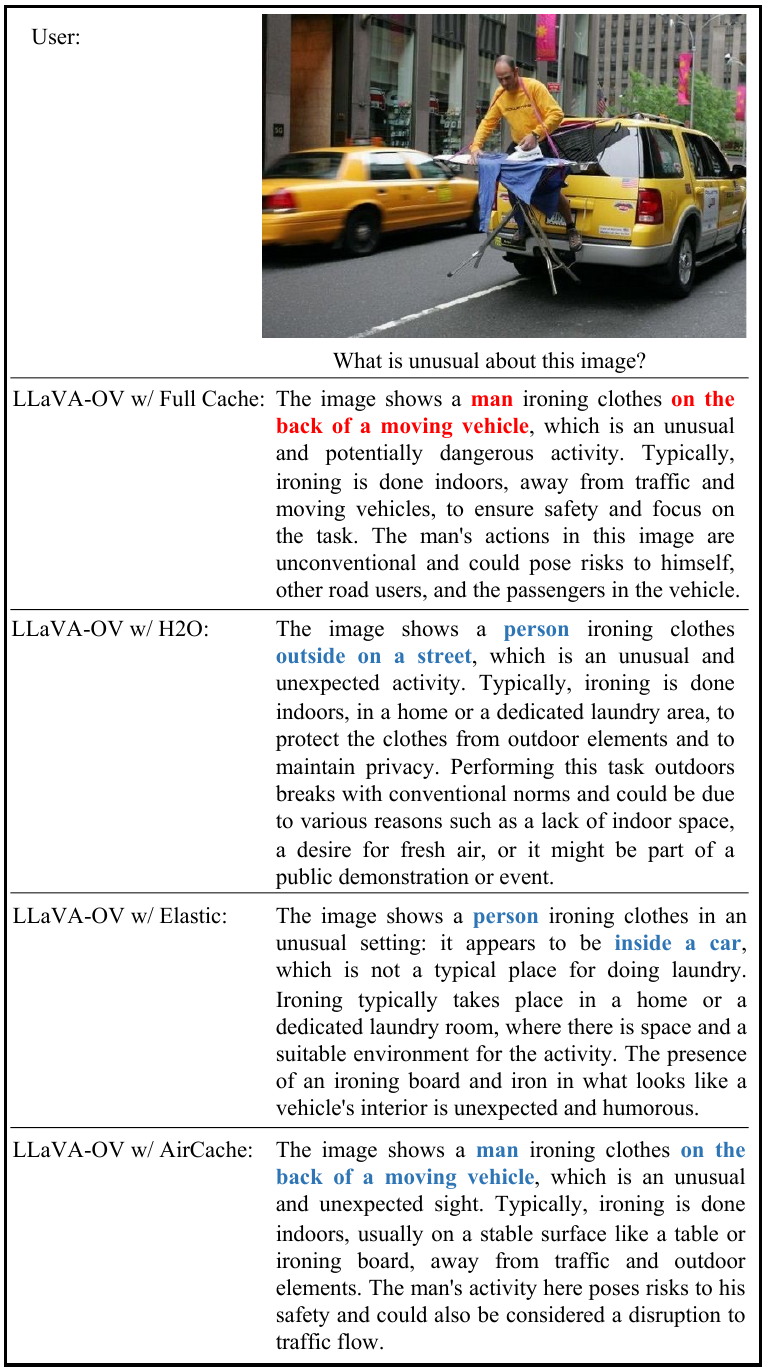}
    \caption{Chat example applying KV cache compression methods on LLAVA-OV-7B \cite{li2024llava}. Important information is highlighted in \textcolor{red}{red} and \textcolor{blue}{blue}.}
    \vspace{-10pt}
    \label{fig10: chat_generation_e4}
\end{figure}

Figures \ref{fig7: chat_generation_e1}, \ref{fig8: chat_generation_e2}, \ref{fig9: chat_generation_e3}, and \ref{fig10: chat_generation_e4} illustrate a comparison of different methods applied to real chat generation while retaining only 1\% of the visual KV cache. It is evident that the answers generated by the proposed method are more accurate.

\end{document}